\documentclass{article} 

\usepackage{jmlr2e}
\usepackage{array} 
\usepackage[utf8]{inputenc} 
\usepackage[T1]{fontenc}    
\usepackage{hyperref}       
\hypersetup{colorlinks,linkcolor=blue,
            citecolor=blue,
            urlcolor=blue,
            linktocpage,
            plainpages=false}
\usepackage{url}            
\usepackage{booktabs}       
\usepackage{amsfonts}       
\usepackage{nicefrac}       
\usepackage{microtype}      
\usepackage{lipsum}
\usepackage{amsmath,amssymb,amsfonts}
\usepackage{bbm}
\usepackage[english]{babel}
\usepackage{graphicx}
\usepackage{caption, subcaption}
\DeclareCaptionLabelFormat{codeexample}{}
\usepackage{xr}
\usepackage{dsfont}
\usepackage[linesnumbered,ruled,vlined]{algorithm2e}
\usepackage{algpseudocode}
\usepackage{natbib} 
\usepackage{thmtools,thm-restate}
\usepackage{wrapfig}
\usepackage{authblk}
\usepackage{tablefootnote}
\usepackage{xcolor}
\usepackage{geometry}
\usepackage{pifont}
\usepackage{enumitem}

\usepackage{listings}
\DeclareFixedFont{\ttb}{T1}{txtt}{bx}{n}{9.5} 
\DeclareFixedFont{\ttm}{T1}{txtt}{m}{n}{9.5}  
\definecolor{codered}{RGB}{255,100,100}
\definecolor{codeblue}{rgb}{0,0,0.6}
\definecolor{codegreen}{rgb}{0,0.6,0}
\definecolor{dark-blue}{rgb}{0.15,0.15,0.4}
\definecolor{codepurple}{rgb}{0.6,0,0.6}
\definecolor{codenumber}{rgb}{0.4,0.8,0}
\definecolor{codebracket}{RGB}{200,200,0}

\newcommand\pythonstyle{\lstset{
    language=Python,
    basicstyle=\fontsize{9.5pt}{10.25pt}\selectfont\ttfamily,,
    otherkeywords={self},             
    keywordstyle=\color{codepurple},
    emph={[1]__init__, dim, None, state_dim, action_space, hidden_dims, learning_rate,
    training_rounds, action_representation_module,  max_number_actions, variance_weighting_coefficient,
    observation_dim, action_dim, history_length, hidden_dim, num_layers,
    },
    emphstyle={[1]\color{codeblue}},
    emph={[2]policy_learner, replay_buffer, safety_module, exploration_module, history_summarizaiton_module},
    emphstyle={[2]\color{codered}},
    stringstyle=\color{codegreen},
    commentstyle=\color{codegreen},
    frame=none,              
    showstringspaces=false,
    breaklines=false,
    numbers=none,
    numbersep=10pt,
    tabsize=2,
    breakatwhitespace=false,
    abovecaptionskip=2ex,
    captionpos=b,
    alsoletter={.},
    literate=*{0}{{{\color{codenumber}0}}}{1}%
             {1}{{{\color{codenumber}1}}}{1}%
             {1e-4}{{{\color{codenumber}1e-4}}}{4}%
             {2}{{{\color{codenumber}2}}}{1}%
             {3}{{{\color{codenumber}3}}}{1}%
             {4}{{{\color{codenumber}4}}}{1}%
             {5}{{{\color{codenumber}5}}}{1}%
             {6}{{{\color{codenumber}6}}}{1}%
             {7}{{{\color{codenumber}7}}}{1}%
             {8}{{{\color{codenumber}8}}}{1}%
             {9}{{{\color{codenumber}9}}}{1}%
}}

\lstnewenvironment{python}[1][]
{
    
    \pythonstyle
    \lstset{#1}
}{}

\newcommand{\pearl}{{\tt Pearl}}
\newcommand{\pearlagent}{{\tt PearlAgent}}

\newcommand{\E}{\mathbb{E}}

\newcommand{\cmark}{\ding{51}}%
\newcommand{\xmark}{\ding{55}}%

\usepackage{lastpage}
\jmlrheading{25}{2024}{1-\pageref{LastPage}}{2/24; Revised
6/24}{8/24}{24-0196}{Zheqing Zhu, Rodrigo de Salvo Braz, Jalaj Bhandari, Daniel R. Jiang, Yi Wan, Yonathan Efroni, Liyuan Wang, Ruiyang Xu, Hongbo Guo, Alex Nikulkov, Dmytro Korenkevych, Urun Dogan, Frank Cheng, Zheng Wu, Wanqiao Xu}

\ShortHeadings{Pearl: A Production-Ready Reinforcement Learning Agent}{Zhu et al.}
\firstpageno{1}

\begin{document}

\title{Pearl: A \textbf{P}roduction-R\textbf{ea}dy \textbf{R}einforcement \textbf{L}earning Agent}

\author{Zheqing Zhu*, Rodrigo de Salvo Braz, Jalaj Bhandari, Daniel R. Jiang, Yi Wan, Yonathan Efroni, Liyuan Wang, Ruiyang Xu, Hongbo Guo, Alex Nikulkov,
Dmytro Korenkevych, Urun Dogan, Frank Cheng, Zheng Wu, Wanqiao Xu
}
\affil[]{Applied Reinforcement Learning Team, AI at Meta \\
*Corresponding author. Please email billzhu@meta.com.} 

%


\editor{Zeyi Wen}

\maketitle

\begin{abstract}
Reinforcement learning (RL) is a versatile framework for optimizing long-term goals. Although many real-world problems can be formalized with RL, learning and deploying a performant RL policy requires a system designed to address several important challenges, including the exploration-exploitation dilemma, partial observability, dynamic action spaces, and safety concerns. 
While the importance of these challenges has been well recognized, existing open-source RL libraries do not explicitly address them.
This paper introduces \pearl, a Production-Ready RL 
software package designed to embrace these challenges in a \emph{modular} way. In addition to presenting benchmarking results, we also highlight examples of \pearl's ongoing industry adoption to demonstrate its advantages for production use cases.
Pearl is open sourced on GitHub at \href{http://github.com/facebookresearch/pearl}{github.com/facebookresearch/pearl} and its official website is \href{http://pearlagent.github.io}{pearlagent.github.io}.
\end{abstract}


\renewcommand{\thefootnote}{\arabic{footnote}}

\vspace{-6pt}
\section{Introduction}
The field of \emph{reinforcement learning} (RL) has achieved significant successes in recent years, including surpassing human-level performance in games \citep{mnih2015human, silver2017mastering}, controlling robots in complex manipulation tasks \citep{peng2018sim, levine2016end, lee2020learning}, optimizing large scale recommender systems \citep{xu2023optimizing}, and fine-tuning of large language models \citep{ouyang2022training}. At the same time, open-source RL libraries, such as Tianshou \citep{tianshou} and Stable-Baselines 3 \citep{stable-baselines3}, have emerged as important tools for reproducible research in RL. However, these existing libraries focus mostly on implementing different algorithms for policy learning, i.e. methods to learn a policy from a stream of data. Even though this is a core feature in RL, there is more to designing an RL solution that can be used for real-world applications and production systems. For example, how should the stream of data be collected? This is known as \emph{exploration}, which is a well-known challenge in RL \citep{sutton2018reinforcement}. Another key question that system designers face when deploying RL solutions is: how should safety concerns be incorporated \citep{dulac2019challenges}?  

In this paper, we introduce {\tt Pearl}, an open-source software package that aims to enable users to easily design practical RL solutions. Beyond just standard policy learning, {\tt Pearl} also allows users to effortlessly incorporate additional features into their RL solutions, including intelligent exploration, safety constraints, risk preferences, state estimation under partial observability, and dynamic action spaces. Many of these features are critical for taking RL from research to deployment and are largely overlooked in existing libraries. To do this, we focus on \textit{agent modularity}, i.e., a modular design of an RL agent that allows users to mix and match various features. {\tt Pearl} is built on native PyTorch, supports GPU-enabled training, adheres to software engineering best practices, and is designed for distributed training, testing, and evaluation. Remaining sections of the paper give details about \pearl's design and features, a comparison with existing open-source RL libraries, and a few examples of \pearl's ongoing adoption in industry applications. We provide examples of using \pearl\, in Appendix\ \ref{sec:pearl_usage}. Results of benchmarking experiments on various testbeds are shown in Appendix\ \ref{sec:benchmark_results}.

\vspace{-10pt}
\section{Pearl Agent Design}
\label{sec: pearl_agent}
\begin{figure}
    \centering
    \includegraphics[width=0.87\textwidth, height=0.33\textwidth]{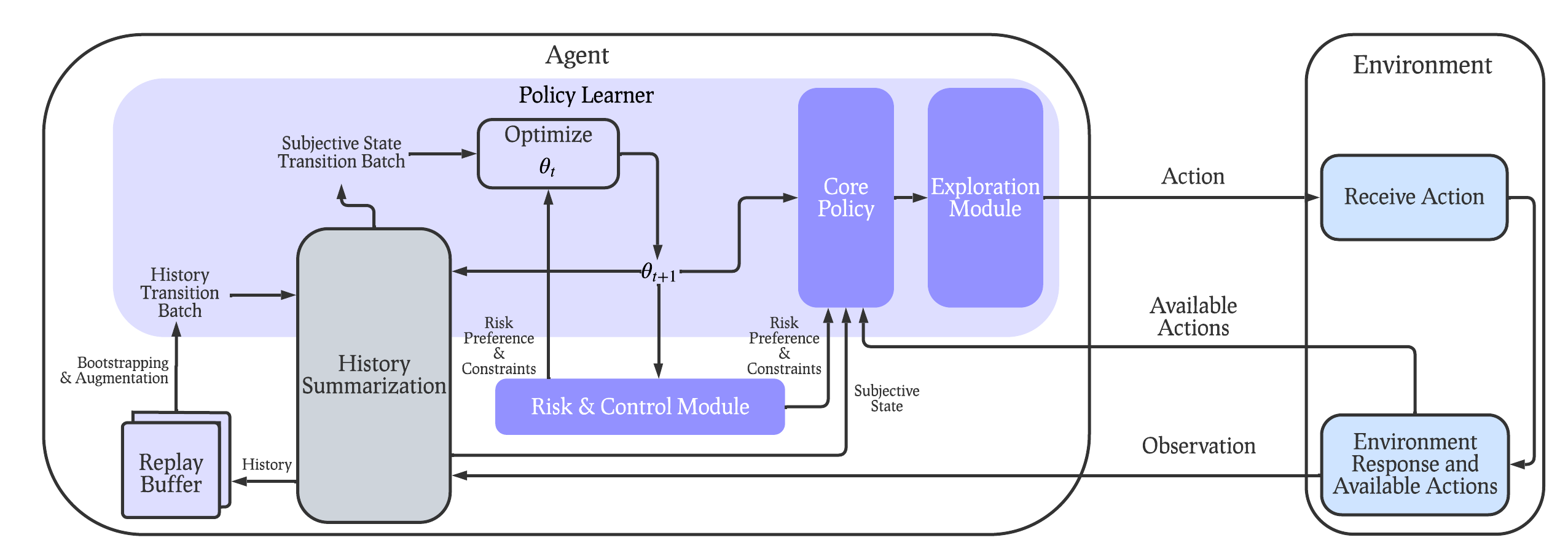}
    \vspace{-6pt}
    \caption{Interface of Pearl Agent}
    \label{fig:agent_interface}
    \vspace{-12pt}
\end{figure}

We give a brief overview of {\tt Pearl}'s design and interface. \pearl\, has four main modules: the \textit{policy learner module}, the \textit{exploration module}, the \textit{history summarization module}, and the \textit{safety module}. These modules implement several key elements that we think are essential for efficient learning in practical sequential decision-making problems. The main contribution of \pearl\, is to support these features in a unified way: Figure \ref{fig:agent_interface} shows how these modules are designed to interact with each other in an online RL setting. Notice that in designing \pearl, we make a clear separation between the RL agent and the environment. In some cases, however, users may only have access to data collected offline, so \pearl\, also offers support for offline RL.\footnote{Learns from a replay buffer with an offline dataset and doesn't interact with the environment during learning.}
Below, we give details of the methods implemented in each module.

\paragraph{Policy learner module:} \pearl\, implements different algorithms for learning a policy for both the contextual bandit and 
the fully sequential RL
problem settings. It is noteworthy that \pearl\, supports policy learning under dynamic action spaces\footnote{An environemnt with dynamic action spaces could feature different action spaces in different states.}, a feature not found in existing libraries. 
\vspace{-2pt}
\begin{enumerate}[leftmargin=*]
    \item[\textbullet] Contextual bandits: Bandit learning methods involve reward modeling and using an exploration module for efficient exploration.
    \pearl\, supports \textit{linear} and \textit{neural bandit} learning along with different exploration modules, as well as the SquareCB method \citep{foster2020beyond}.
    \vspace{-2pt}
    \item[\textbullet] Value-based methods: \pearl\, supports DQN \citep{mnih2015human}, Double DQN \citep{van2016deep}, Dueling DQN \citep{wang2016dueling}, Deep SARSA \citep{rummery1994line} and Bootstrapped DQN \citep{osband2016deep}---a DQN variant with ensemble-based deep exploration.
    \item[\textbullet] \looseness-1 Actor-critic methods: SAC \citep{haarnoja2018soft}, DDPG \citep{silver2014deterministic}, TD3 \citep{fujimoto2018addressing}, PPO \citep{schulman2017proximal}, and REINFORCE \citep{sutton1999policy}.
    \vspace{-2pt}
    \item[\textbullet] Offline RL methods: CQL \citep{kumar2020conservative}, IQL \citep{kostrikov2021offline} and TD3BC \citep{fujimoto2021minimalist}. 
    \vspace{-2pt}
    \item[\textbullet] Distributional RL methods: Quantile Regression DQN (QRDQN) \citep{dabney2018distributional}.
\end{enumerate}
\vspace{-6pt}

\paragraph{Exploration module:} \looseness-1 
One of the main challenges in RL is to determine how the agent should collect data through intelligent exploration for faster learning. We design the {\tt exploration\char`_module}\, in \pearl\,in a way such that it supports implementation of different intelligent exploration algorithms for contextual bandit and fully sequential RL settings.
Popular upper confidence bound (UCB) and Thompson sampling (TS) methods for contextual bandit problems such as LinUCB \citep{li2010contextual}, Neural-LinUCB  \citep{xu2021neural}, and LinTS \citep{agrawal2013thompson}, along with the recent SquareCB method by \cite{foster2020beyond}, are included in \pearl\,.
For the RL setting, we implement the idea of deep exploration by ensemble sampling \citep{osband2016deep}, which performs temporally consistent exploration using approximate posterior samples of the optimal value function. Naive exploration strategies that are commonly implemented in existing libraries, such as $\epsilon$-greedy exploration \citep{sutton2018reinforcement}, Gaussian exploration for continuous action spaces \citep{lillicrap2015continuous}, and Boltzmann exploration \citep{cesa2017boltzmann} are also implemented in \pearl\,. 

\paragraph{Safety module:} Safe RL is another key hurdle in deploying RL to solve real world problems \citep{garcia2015comprehensive, dulac2019challenges}. \pearl\, provides three features to enable safe learning. The {\tt filter\char`_action} method in \pearl's safety module allows users to specify a \emph{safe action space} for every environment interaction. In addition, the {\tt risk\char`_sensitive\char`_safety\char`_module} facilitates risk-sensitive learning using distributional policy learners, which approximate the distribution of cumulative rewards. \pearl\,then implements various methods to compute the Q-value/value function under different risk metrics from this distribution. \pearl\,is also designed to enable learning in \textit{constrained MDPs}. This is done through the {\tt reward\char`_constrained\char`_safety\char`_module} which allows users to specify constraints on cumulative costs\footnote{Users can specify both the per-step costs as well as the threshold (upper bound) on long run costs.}, in addition to the reward maximization objective. We implement the reward constrained policy optimization (RCPO) method of \cite{tessler2018reward} in this module, which is compatible with different policy learners and general cost constraints.

\paragraph{History summarization module:} Applications of RL which require learning under partial observability of states are ubiquitous \citep{levinson2011towards, hauskrecht2000planning}. To facilitate learning in partially observable settings, we equip \pearl\, with a history summarization module that implements two functionalities. First, it tracks the history (and updates it) after every environment interaction. Instead of raw observations, histories are stored in the interaction tuple at every time step. Second, during policy learning (and action selection), this module provides different ways to summarize the history into a state representation. \pearl\, currently supports using history via stacking of past state-action pairs as well as using a long-short-term-memory (LSTM) neural network \citep{hochreiter1997long} to learn a state representation. The history summarization module and the policy can be (jointly) trained in an end-to-end fashion using a \emph{single} backward pass.

\vspace{-8pt}
\section{Comparison to Existing Libraries}
\vspace{-12pt}
\begin{table}[ht]
\caption{Comparing features of Pearl to existing popular open-source RL libraries}
\vspace{-7pt}
\centering
\begin{tabular}{c|c|c|c|c|c|c|c }
    Features & ReAgent 
     & RLLib & SB3 & Tianshou & CleanRL & TorchRL & Pearl \\
    \hline
    Agent modularity & \xmark &\xmark & \xmark & \xmark &\xmark & \xmark &  \cmark\\
    Intelligent exploration & \xmark &\xmark &  \xmark & \cmark & \xmark & \xmark & \cmark\\
    Safe learning & \xmark & \xmark & \xmark  & $\circ$\tablefootnote{\label{note:safety}Even though Tianshou and CleanRL have implementations of quantile regression DQN and/or C51, these are more like standalone algorithm implementations which do not implement generic risk sensitive learning. In addition, none of the existing libraries implement policy learning with constraints. 
    } & $\circ$\footnotemark[4] &  \xmark & \cmark\\
    History summarization & \xmark &  \cmark & \xmark & \xmark & \xmark  & \xmark &\cmark \\
    Contextual bandits  & \cmark &$\circ$\tablefootnote{Only supports linear bandit learning algorithms.} & \xmark & \xmark & \xmark & \xmark & \cmark \\
    Offline RL & \cmark & \cmark & \cmark & \cmark & \xmark  & \cmark &  \cmark \\
    Dynamic action spaces & $\circ$\tablefootnote{ Only support for value based methods is provided.} &\xmark & \xmark & \xmark &\xmark & \xmark & \cmark \\
\end{tabular}
\label{tab:comparison}
\vspace{-7pt}
\end{table}

\label{sec: lit_review}
To illustrate \pearl's innovations, we compare its functionalities to those of other popular RL libraries like ReAgent~\citep{gauci2018horizon}, RLLib~\citep{liang2018rllib}, StableBaselines3~\citep{stable-baselines3}, Tianshou~\citep{tianshou}, CleanRL~\citep{huang2022cleanrl}, and TorchRL ~\citep{bou2023torchrl}. Fundamentally, the main motivation of these libraries is to facilitate research and reproducible benchmarking of existing RL algorithms. On the other hand, \pearl\, is designed to implement several key capabilities that are crucial for developing an RL-based system that can be deployed in practice. 


Table~\ref{tab:comparison} compares several of these features.\footnote{We do not include policy learning in the comparison as it is a core component in all libraries.} As shown, all existing libraries provide little to no support for safe learning. With the exception of ReAgent, none of the other libraries support dynamic action spaces either.
Support for intelligent exploration methods and for the ability to learn in partially observable environments is also sparse amongst the existing libraries. Finally, \pearl\, is among a handful of other libraries (besides ReAgent and to some extent RLLib) to support learning in both contextual bandit and fully sequential RL settings on the same platform -- this is important as many large-scale industry applications are often modeled as contextual bandits.

We emphasize that the main innovation behind \pearl\, is the ability to combine the aforementioned features in a unified way. Users do not need to write code from scratch to test out different combinations of features. For example, implementations of policy learning algorithms in other libraries often come paired with a fixed exploration strategy (e.g., DQN with $\epsilon$-greedy exploration). However, \pearl\, enables the possibility to mix and match different policy learning and exploration methods, which can hopefully lead to more performant RL solutions. Modular design of the RL agent in \pearl\, makes this and other combinations of features among exploration, safe learning, history summarization, dynamic action spaces, and offline RL possible. It also allows both academic and industry researchers to quickly add new modules to efficiently test new ideas. 


\vspace{-10pt}
\section{Adoption of Pearl in Industry Applications}
\label{sec:product_adoptions}
We are currently integrating \pearl\,to enable RL-based solutions in large scale production systems, to solve three real-world problems. Table \ref{tab:industry} below highlights some unique capabilities of \pearl, enabled by a focus on a modular agent design, which can serve different product requirements in each case.


\looseness-2 \paragraph{Auction-based recommender systems:} We are using \pearl\,to implement our recent algorithm design in \citep{xu2023optimizing} to use RL to optimize for long-term value in an auction-based recommender system.
This solution requires estimating the Q-value function of the auction-based policy using data collected online, for which we use \pearl's implementation of on-policy offline temporal-difference learning. Moreover, as a unique set of recommendations is available at each time step, \pearl's ability to work with dynamic action spaces is critical for our implementation. \pearl's modular design easily supports the integration of large-scale neural networks, which in this case are used for estimating Q-values for user-item interactions.

\paragraph{Ads auction bidding:} 
We are also using \pearl\,to implement an RL system for real-time auction bidding in advertising markets, where the task is to make effective bids subject to advertiser budget constraints. For this, we follow a solution outlined in our recent work in \cite{korenkevych2023offline}, which requires collecting data (with some exploration) and then training an agent using offline RL. We leverage the offline training pipeline and the Gaussian exploration module in \pearl\,for our implementation. Furthermore, we are using \pearl\,to test two extensions of the work in \cite{korenkevych2023offline}. The first one uses \pearl's constrained policy optimization submodule to account for advertiser's budget constraints. The other extension uses \pearl's history summarization module to encode interaction history into the agent's state representation. This is useful because advertising markets are partially observable (only the winning bid is observed).

\begin{table}[ht]
\vspace{-6pt}
\caption{\pearl\, incorporates features that are critical for real-world adoption}
\vspace{-6pt}
\centering
\begin{tabular}{c|c|c|c}
    \pearl\, Features & Auction RecSys & Ads Auction Bidding & Creative Selection\\
    \hline
    Online exploration & & \cmark & \cmark \\
    Safe learning &  & \cmark & \\
    History summarization &  & \cmark &  \\
    Contextual bandits  & &  & \cmark \\
    Offline RL & \cmark & \cmark &  \\
    Dynamic action spaces &\cmark &  & \cmark\\
\end{tabular}
\label{tab:industry}
\end{table}

\vspace{-10pt}
\paragraph{Creative selection:} 
In this project, \pearl\,is being used by our team to implement a contextual bandit solution for the problem of creative selection. Creative selection is commonplace in social media and online advertising and refers to the problem to choosing a combination of ad components, such as image and text, along with their placement, font, or color, to present to users. We cast this as a contextual bandit problem and are using \pearl\,to implement a solution with a neural bandit policy learner along with neural linear UCB exploration, to learn user preferences for different creatives. Creative selection is another example where the action space changes dynamically, as each piece of content can be shown using a corresponding set of available creatives. Again, \pearl's support for learning with dynamic action spaces greatly simplifies our implementation.

\newpage
\appendix
\section{Using Pearl}
\label{sec:pearl_usage}

We show some examples of how to instantiate an RL agent in \pearl, which is done using the \pearlagent\, class. In Code Example \ref{codex:Abstractions:ImplementationExamples:Example1} below, we create a simple agent which uses Deep Q-learning (DQN) for policy optimization and $\epsilon$-greedy exploration. It is worth emphasizing that the {\tt exploration\char`_module} in \pearl\, is passed as an input to the {\tt policy\char`_learner}. This design choice is important to allow implementation of intelligent exploration algorithms. 

\begin{python}[
caption=A \pearlagent\, with DQN as the policy learner and $\epsilon$-greedy exploration.,
label=codex:Abstractions:ImplementationExamples:Example1,
float=ht!
]
env = GymEnvironment("CartPole-v1")
assert isinstance(env.action_space, DiscreteActionSpace)
num_actions: int = env.action_space.n

agent = PearlAgent(
    policy_learner=DeepQLearning(
        state_dim=env.observation_space.shape[0],
        action_space=env.action_space,
        hidden_dims=[64, 64],
        learning_rate=1e-4,
        exploration_module=EGreedyExploration(0.10),
        training_rounds=5,  # number of gradient updates per learning step
        action_representation_module=OneHotActionTensorRepresentationModule(
            max_number_actions=num_actions
        ),
    ),
    replay_buffer=FIFOOffPolicyReplayBuffer(10000),
)
\end{python}

\begin{python}[
caption={A \pearlagent\, with a distributional policy learner (QRDQN) and the mean variance safety module, which maps the estimated distribution over long-run rewards to Q-values.},
label=codex:Abstractions:ImplementationExamples:Example2,
float=ht!
]
env = GymEnvironment("CartPole-v1")
assert isinstance(env.action_space, DiscreteActionSpace)
num_actions: int = env.action_space.n
agent = PearlAgent(
    policy_learner=QuantileRegressionDeepQLearning(
        state_dim=env.observation_space.shape[0],
        action_space=env.action_space,
        hidden_dims=[64, 64],
        learning_rate=1e-4,
        exploration_module=EGreedyExploration(0.10),
        training_rounds=5,
        action_representation_module=OneHotActionTensorRepresentationModule(
            max_number_actions=num_actions
        ),
    ),
    safety_module=QuantileNetworkMeanVarianceSafetyModule(
        variance_weighting_coefficient=0.2
    ),
    replay_buffer=FIFOOffPolicyReplayBuffer(10000),
)

\end{python}

Code Example \ref{codex:Abstractions:ImplementationExamples:Example2} illustrates how \pearl\, can be used for risk-sensitive learning. In this example, QRDQN (a distributional RL method) is used for policy learning, along with a mean-variance risk-sensitive safety module. Note that risk-sensitive learning necessitates the use of both a distributional policy learner and a risk sensitive safety module---the former estimates a distribution over long run discounted rewards and uses the risk measure implemented in the latter to compute the Q-values (which are used for policy learning). Compatibility checks are implemented in \pearl\, to raise errors when users inadvertently try to use a combination of modules incompatible with each other.\footnote{For example, specifying {\tt DeepQLearning} as the {\tt policy\char`_learner} and {\tt QuantileNetworkMeanVarianceSafetyModule} as the {\tt safety\char`_module} when instantiating a {\tt PearlAgent} will raise an error.} We also provide defaults to aid new users who want to use \pearl, but are not familiar with the details. For brevity, we do not show more code examples and refer users to \pearl's documentation and tutorials on the official website for more illustrations.

Both code examples \ref{codex:Abstractions:ImplementationExamples:Example1} and \ref{codex:Abstractions:ImplementationExamples:Example2} show that a {\tt replay\char`_buffer} also needs to be specified when instantiating a \pearlagent. In \pearl, different replay buffers are implemented for learning in various problem settings. For example, first-in-first-out (FIFO) on-policy and off-policy buffers enable learning in on-policy and off-policy settings respectively.  

We end this section with sample code on the agent-environment interface in \pearl\,as shown below in Code Example \ref{codex:Abstractions:ImplementationExamples:Example3}.
The clear separation in \pearl\,between the agent and the environment is worth emphasizing: \pearl\,users do not have to explicitly specify an environment. This is a useful feature for practical applications, as \pearl\,users can build agents to directly interact with the real-world, for example, humans, with API wrappers\footnote{\pearlagent\,requires an {\tt ActionResult} object to observe and store data.} to enable learning. For instance, consider designing a chatbot agent in \pearl\, that can output text responses to the user as an action. The agent could then receive the next text query as an observation, wrapped according to the \pearlagent\,API. Here, online interactions can be enabled without having to specify {\tt reset()} and {\tt step()} methods typically required by an {\tt Environment} class instance. Therefore, \pearl\, is not limited to being used with simulation environments (e.g., Open AI Gym or MuJoCo).


\begin{python}[
caption=An illustration of how users can use an RL agent in \pearl\, to interact with an environment and learn online. Note that the input environment is only required for implementing the {\tt env.reset()} and {\tt env.step()} functions.,
label=codex:Abstractions:ImplementationExamples:Example3,
float=ht!
]

def online_learning_example(
    agent: PearlAgent,
    env: Environment,
    num_episodes: int,
    learn: bool = True,
    exploit: bool = True,
    learn_after_episode: bool = False,
) -> List[float]:
    """
    Runs the specified number of episodes and returns a list with
    episodic returns.

    Args:
        agent (Agent): the agent.
        env (Environment): the environment.
        num_episodes (int): number of episodes to run.
        learn (bool, optional ): Runs `agent.learn()` after every step.
        exploit (bool, optional ): asks the agent to only exploit.
            Defaults to False. If set to True , then exploration module
            of the agent is used to explore.
        learn_after_episode (bool, optional): learn only at the end of each episode.
    Returns:
        List[float]: returns of all episodes.
    """
    returns = []
    for i in range(num_episodes):
        observation, action_space = env.reset()
        agent.reset(observation, action_space)
        cum_reward = 0
        done = False
        while not done:
            action = agent.act(exploit=exploit)
            action_result = env.step(action)
            cum_reward += action_result.reward
            agent.observe(action_result)
            # to learn after every environment interaction
            if learn and not learn_after_episode:
                agent.learn()
            done = action_result.done
        # to learn only at the end of the episode
        if learn and learn_after_episode:
            agent.learn()
        returns.append(cum_reward)
        print(f"Episode {i}: return = {cum_reward}")
    return returns

\end{python}

\newpage
\newpage

\subsection{Architecture, Modularity and API design}
\label{subsec:architecture}

As explained in the main text, \pearl\, consists of four main types of modules: policy learner, exploration module, history summarization module, and safety module. In this section we provide more details on \pearl's architecture and on how these modules interact.

The policy learner, history summarization module, and safety module are top-level components, that is, they are present at the top-level \pearl\, agent level. The exploration module, on the other hand, is a component of the policy learner. The flow of information occurs in the following manner:

\begin{itemize}
    \item When an observation is provided to the \pearl\, agent, the history summarization module transforms it into a \emph{state}, which is the internal representation used by the other modules. Any history summarization module can do this, including the trivial method of using the observation as the state, or more elaborately generating state representations that take previous observations into account. The simplest version of the latter is to update a sliding window of $n$ stacked past observations, while a more sophisticated module uses LSTMs to produce a state that encodes the main characteristics of the agent's history. Examples of agents using these modules are shown in Code Examples \ref{codex:stacked-history-example} and \ref{codex:lstm-history-example}.

\begin{python}[
caption=A \pearlagent\ defined to use the stacking history summarization module.,
label=codex:stacked-history-example,
float=ht!
]
agent = PearlAgent(
    policy_learner=DeepQLearning(
        state_dim=observation_dim * 3,  # *3 due to history summarization
        action_space=env.action_space,
        hidden_dims=[64, 64],
        training_rounds=20,
        exploration_module=EGreedyExploration(epsilon=0.05),
    ),
    history_summarization_module=StackingHistorySummarizationModule(
        observation_dim=observation_dim, 
        history_length=3
    ),
    ...
)
\end{python}

\begin{python}[
caption=A \pearlagent\ defined to use the LSTM history summarization module.,
label=codex:lstm-history-example,
float=ht!
]
agent = PearlAgent(
    policy_learner=DeepQLearning(
        state_dim=observation_dim,
        action_space=env.action_space,
        hidden_dims=[64, 64],
        exploration_module=EGreedyExploration(epsilon=0.05),
    ),
    history_summarization_module=LSTMHistorySummarizationModule(
        observation_dim=observation_dim,
        action_dim=action_dim,
        hidden_dim=128,
        history_length=8,
    ),
    ...
)
\end{python}

    \item In the process of deciding which action to take, safety issues come into play: subsets of actions may be discouraged (through a soft measure of \emph{cost} or via risk preferences specified by the user) or outright forbidden. Support for these situations is provided by safety modules. Currently we provide a reward constrained safety module \citep{tessler2018reward} and a quantile network mean variance safety module. The latter learns a distribution of returns for each state-action pair; and computes the Q-values under the risk measure specified by the user. An example of \pearlagent\, defined to use the quantile network mean variance safety module is shown in Code Example \ref{codex:quantile-safety-module-example}.

\begin{python}[
caption=A \pearlagent\ defined to use the quantile network mean variable safety module (which must be used jointly with quantile regression deep Q-learning policy learner).,
label=codex:quantile-safety-module-example,
float=ht!
]
agent = PearlAgent(
    policy_learner=QuantileRegressionDeepQLearning(
        state_dim=observation_dim,
        action_space=action_space,
        hidden_dims=[64, 64],
        exploration_module=EGreedyExploration(epsilon=0.05),
        safety_module=QuantileNetworkMeanVarianceSafetyModule(
            variance_weighting_coefficient=0.2,
        ),
        ...
    )
)
\end{python}

    \item The policy learner then uses the state and the space of available actions to decide on an action. Besides the RL algorithm it implements, the policy learner also makes a decision based on whether it is in exploratory mode or not, and what exploration strategy it is using, which is defined by the specific exploration module provided by the user. 
    \begin{itemize}
        \item Some exploration strategies, especially those based on posterior sampling \citep{osband2016deep} require access to various intermediate pieces of information computed by the policy learner, such as the value of each action or uncertainty estimates. The need for this information is why the exploration module must be a component of the policy learner (as opposed to a post-processing module external to the policy learner that only gets to see the action chosen by the former). The policy learner provides the required information to the exploration module through a special exploration module API, which then responds with the appropriate action to be taken given the current exploration strategy. Note that the policy learner can be combined with different types of exploration modules since it does not participate in the exploration calculus; it merely provides the information required by the exploration module that was chosen by the user, which then makes its own decision. This makes it easy for users to use or write novel exploration modules without the need to modify the policy learner as a result. Code Example \ref{codex:quantile-safety-module-example} shows a \pearlagent\, defined with an $\epsilon$-greedy exploration module (as part of its policy learner).
    \end{itemize}
\end{itemize}

\subsection{Handling Errors and Edge Cases}
\pearl\, checks inputs and specifications and raises exceptions when they are invalid. The exceptions can then be caught and dealt with in multiple ways, such as by logging a message or raising an error.

An example of particularly useful error checking by \pearl\, is the compatibility check done by \pearlagent\, on its modules since some modules can only be used in conjunction with other modules. For example, the quantile network mean variance safety module discussed in section \ref{subsec:architecture} can only be used in conjunction with a distributional policy learner, and if the user mistakenly uses it with another type of policy learner, this will be indicated as soon as the \pearlagent\, is instantiated.

Other examples, among many, of error handling include checking the compatibility of batch sizes and replay buffer sizes, of types of actions (discrete or continuous) and algorithms handling one type or the other, and the misuse of a value network (state only) when a Q-value network (state-action pairs) is required.

\section{Benchmark Results}
\label{sec:benchmark_results}
In this section, we provide benchmark results for \pearl. We tested \pearl\, in a variety of problem settings, like discrete and continuous control problems, offline learning, and contextual bandit problems. To this end, we evaluated \pearl\ in several commonly used tasks from OpenAI Gym \citep{brockman2016openai} and modified versions of some of these tasks. Our experiments were carefully done to test combinations of all four modules (policy learning, exploration, safe learning and history summarization). In many cases, we design a simple environment ourselves to emphasize challenges in RL such as partial observability, sparsity of rewards, risk sensitive learning, learning with constraints, and dynamic action spaces. These help elucidate \pearl's main motivation: the importance of considering various aspects of RL (beyond just using popular policy optimization algorithms) in designing a system that can deal with challenges of real-world applications. 

We first present speed and memory usage for Pearl and compare it to another high-quality open-source library \ref{sec: Speed and Memory Usage}. We then present results for benchmarking policy learning methods in \pearl\, for RL and contextual bandit problems in subsections \ref{subsec:RL_benchmarks} and \ref{subsec:cb_benchmarks}. Subsection \ref{subsec:versatility_benchmarks} tests the other three modules in \pearl, along with \pearl's ability to learn in problems with dynamic action spaces.


\subsection{Speed and Memory Usage}\label{sec: Speed and Memory Usage}

Before presenting our benchmark results, we give a preliminary comparison of Pearl's speed and memory usage against CleanRL \citep{huang2022cleanrl}. CleanRL offers single-file, non-modularized implementations of algorithms, focusing on the most stripped-down versions. It does not address additional challenges such as safety and dynamic action spaces. This simple design allows CleanRL, in theory, to achieve fast speeds and low memory usage. Furthermore, CleanRL has well-documented performance on various testbeds, demonstrating comparable effectiveness to other RL implementations, as shown in https://docs.cleanrl.dev/.

We evaluated the Soft Actor-Critic algorithm on HalfCheetah-v4, a test problem from the Mujoco suite with a continuous action space. Using identical parameters for both Pearl and CleanRL implementations, we measured the time and memory usage at 10,000 environment steps. Detailed parameter settings are provided in Section \ref{sec: Continuous control}. The experiments were conducted on an Intel(R) Xeon(R) Platinum 8339HC CPU (1.80GHz) with an Nvidia A100 GPU.

At the 10,000-step mark, Pearl required 145 seconds and consumed 1.0 GB of memory, in contrast to CleanRL, which took 123 seconds and utilized 4.2 GB of memory. CleanRL's higher memory consumption is attributed to its method of initializing a replay buffer filled with a zero vector sized to the buffer's capacity. In contrast, Pearl dynamically allocates memory as needed, up to the buffer limit. By the end of the training period, when the replay buffer was full, CleanRL's memory usage increased to 4.4 GB, whereas Pearl's usage rose to 5.7 GB. The slower performance of Pearl compared to CleanRL can be linked to Pearl's modular design, which is tailored to address specific challenges such as safety and dynamic action spaces. Conversely, CleanRL employs a simpler design that does not tackle these issues.

The increase memory usage of Pearl as compared to CleanRL can be attributed to Pearl's replay buffer design which requires storing a few more attributes such as the available action spaces for the current and next state, cost associated with a state-action pair (if any) etc. This design is necessary to enable different functionalities in Pearl -- for example, handling of dynamic action spaces, learning under cost constraints using the reward constrained safety module etc. As CleanRL only provides specific and limited implementations of policy learning algorithms, its replay buffer design is simpler.

\subsection{Reinforcement Learning Benchmarks}\label{subsec:RL_benchmarks}
\begin{figure}[!htb]
\begin{center}
    \minipage{0.50\textwidth}
    \includegraphics[width=\linewidth]{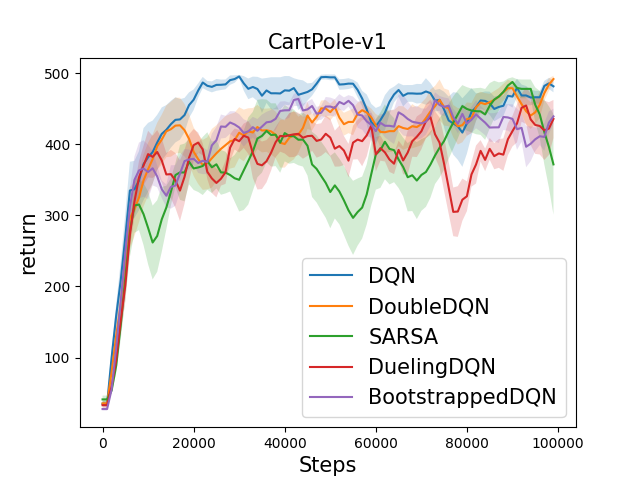}
\endminipage
\minipage{0.50\textwidth}
    \includegraphics[width=\linewidth]{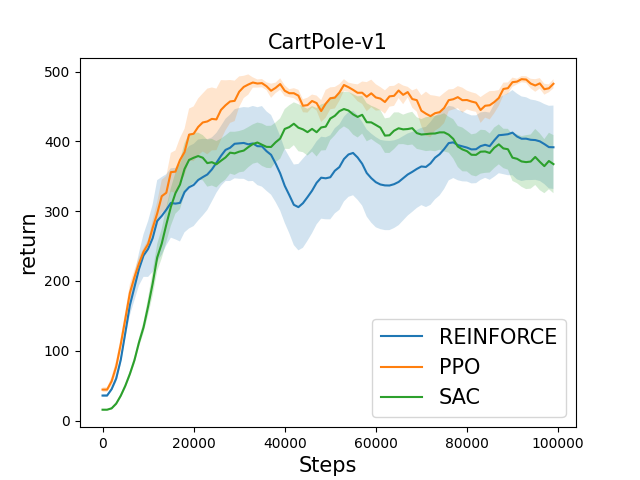}
\endminipage
\end{center}
\caption{Training returns for discrete control methods on the CartPole task. The left and right panels show returns for value- and policy-based methods, respectively. }
\label{fig: discrete control tasks}
\end{figure}

\subsubsection{Discrete control} 
We first benchmark discrete control methods in \pearl\, on Cartpole, a classic reinforcement learning task from 
OpenAI Gym. 
We give details of our experimental setup below. Throughout our experiments, we set the discount factor to be $0.99$.

\paragraph{Experiment Setup for Value-Based Discrete Control Methods.}
We tested \pearl's implementation of value-based discrete control methods including DQN, DoubleDQN, SARSA, DuelingDQN and BootstrappedDQN. For exploration, we used the $\epsilon$-greedy exploration module in \pearl\, with $\epsilon = 0.1$ and set the mini-batch size to be 32. All methods used AdamW optimizer with a learning rate of $10^{-3}$, and updated the target network every $10$ steps. The step size of this update was $0.1$ for SARSA and $1.0$ for other methods. DQN, DoubleDQN, DuelingDQN, and BootstrappedDQN all used \pearl's first-in-first-out (FIFO) replay buffer of size $50,000$, and random sampling for mini-batch updates. SARSA updates its parameters at the end of each episode using all the interaction tuples collected within that episode. Regarding neural network architecture, DQN, DoubleDQN, and SARSA used a fully-connected neuron network (NN) with two hidden layers of size $[64 \times 64]$ to approximate the Q-value function. For Bootstrapped DQN, we used an ensemble of $10$ such NNs to approximate the posterior distribution of the Q-value function.


\paragraph{Experiment Setup for Policy-Based Discrete Control Methods.}
We also tested several of \pearl's policy-based discrete control methods. The tested methods are REINFORCE, proximal policy optimization (PPO) and the discrete control version of soft actor critic (SAC). The policy networks for all three methods were parameterized by a fully-connected NN with two hidden layers of size $[64 \times 64]$ and were updated using the AdamW optimizer with a learning rate of $10^{-4}$. In addition, for PPO we used a critic with the same architecture, a fully-connected NN with two hidden layers of size $[64 \times 64]$. For Discrete SAC, we used a twin critic, which comprised of two fully connected NNs, each with two $[64 \times 64]$ hidden layers. For both PPO and Discrete SAC, these critics were updated by the AdamW optimizer with a learning rate of $10^{-4}$. Our Discrete SAC implementation additionally used a target network for the critic, which was updated using \textit{soft-updates} with a step-size of $0.005$.

Again, both REINFORCE and PPO make updates at the end of each episode, using only data collected within that episode -- for this, we used the episodic replay buffer in \pearl. Discrete SAC, on the other hand, uses mini-batch updates every step -- for this, we used a 50000-sized FIFO replay buffer with a mini-batch size of 32. The clipping lower and upper thresholds for importance sampling ratio in PPO were chosen to be $0.9$ and $1.1$, respectively. 

For all three methods, we used \pearl's propensity exploration module for exploration, which basically means taking actions sampled from their learned policies. The Discrete SAC method uses entropy regularization to improve its exploration ability -- we set the entropy regularization coefficient in SAC to be $0.1$ for this experiment.


\paragraph{Results.}
We plot learning curves of each RL agent as described above in Figure\ \ref{fig: discrete control tasks}. Here, the $x$-axis shows the number of environment steps and the $y$-axis shows episodic returns averaged over past $5000$ steps. Each experiment was performed with $5$ different random seeds. Figure \ref{fig: discrete control tasks} also plots the $\pm1$ standard error across different runs. We note that these results are only meant to serve as a sanity check for combination of policy learning and exploration modules commonly found in implementations of other libraries -- we checked only for stable, consistent learning of our implementations rather than searching for the best training runs with optimal hyperparameter choices.


\subsubsection{Continuous control}\label{sec: Continuous control}
We also benchmarked \pearl\, implementations of three popular actor-critic methods: Continuous soft-actor critic (SAC), Discrete deterministic policy gradients (DDPG) and Twin delayed deep deterministic policy gradients (TD3), on four continuous control tasks in four MuJoCo games \citep{todorov2012mujoco,brockman2016openai}. For all experiments, we set $\gamma=0.99$.

\paragraph{Experiment Setup for Continuous Control Methods.}
All three methods use an actor and a critic. We parameterized both by a fully connected NN with two hidden layers of size $[256 \times 256]$. Both actor and critic networks were trained using mini-batch updates of size 256, drawn from a FIFO replay buffer of size 100,000. DDPG and TD3 are algorithms with deterministic actors. For exploration, we used Gaussian perturbations with a fixed small standard deviation which was set to be 0.1. For SAC, we set the entropy regularization coefficient to be $0.25$.


All three methods used the AdamW optimizer to update both the actor and the critic. For SAC, the learning rates for the actor and the critic were $3 \times 10^{-4}$ and $5 \times 10^{-4}$ respectively. For DDPG and TD3, the learning rates for the actor and the critic were $10^{-3}$ and $3 \times 10^{-4}$ respectively. All three methods used a target network for the critic, with an soft update rate of $0.005$. Additionally, DDPG and TD3 used a target network for the actor networks as well with a soft update rate of $0.005$. 

The results, averaged over five different random seeds, are shown below in Figure \ref{fig:continuous_control_benchmarks}. As before, the $x$-axis shows the number of environment steps and the $y$-axis shows running average of episodic returns over 5000 steps. Again, our goal here was to check for consistent learning rather than searching for the best hyperparameters. Plots in Figure \ref{fig:continuous_control_benchmarks} clearly validate implementations of these actor-critic methods in \pearl.

\begin{figure}[!htb]
\minipage{0.50\textwidth}
    \includegraphics[width=\linewidth]{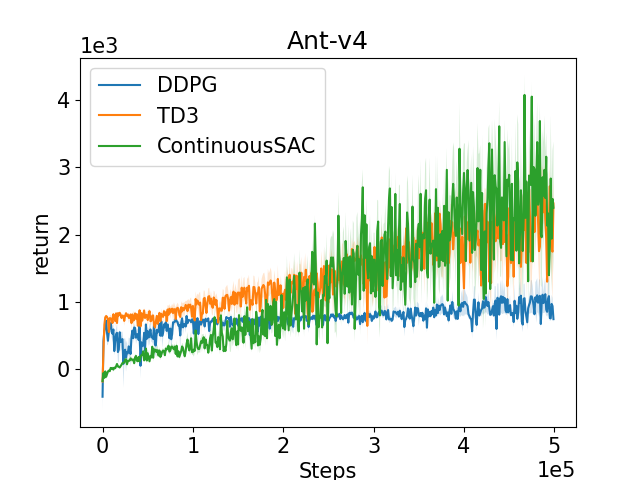}
\endminipage\hfill
\minipage{0.50\textwidth}
    \includegraphics[width=\linewidth]{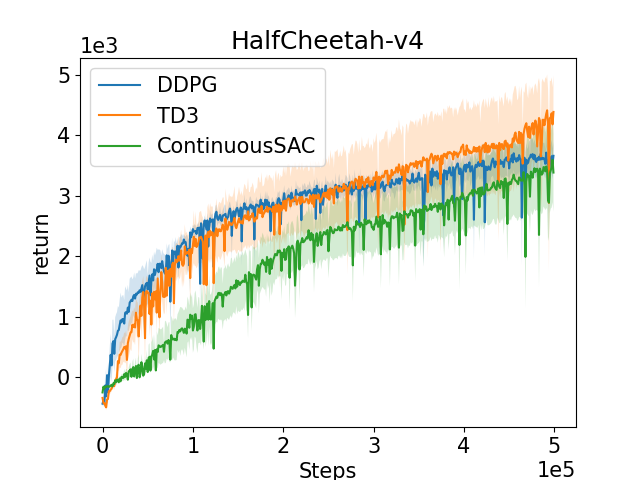}
\endminipage\hfill
\minipage{0.50\textwidth}%
    \includegraphics[width=\linewidth]{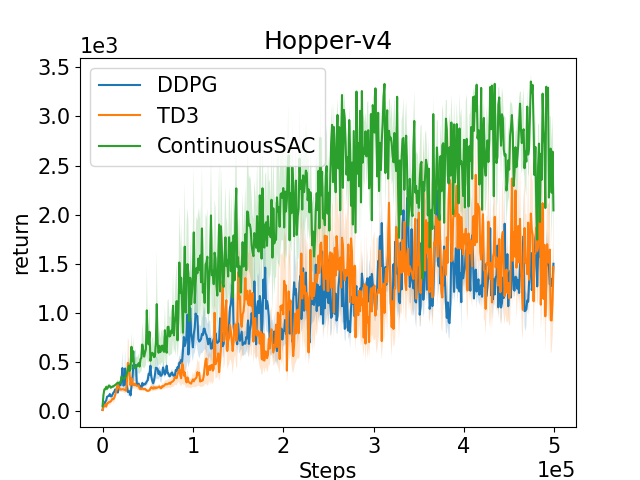}
\endminipage
\minipage{0.50\textwidth}%
    \includegraphics[width=\linewidth]{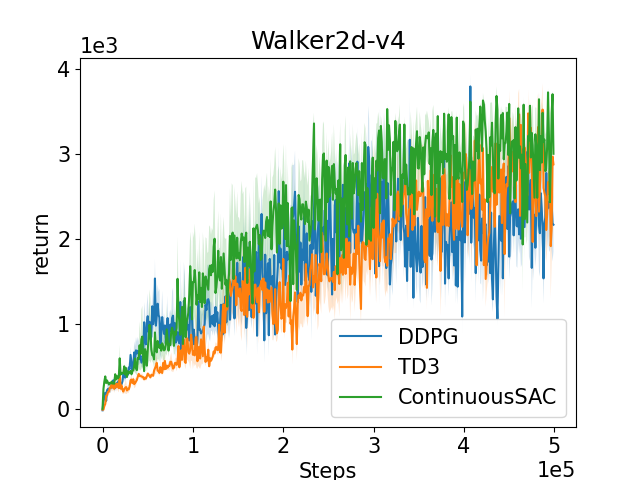}
\endminipage
\caption{Training returns for Continuous SAC, DDPG and TD3 on four continuous control tasks in MuJoCo.}
\label{fig:continuous_control_benchmarks}
\end{figure}

\paragraph{Offline Learning Methods.}
We also tested Implicit Q-learning (IQL) \citep{kostrikov2021offline}, an offline algorithm implemented in \pearl\, on continuous control MuJoCo tasks with offline data. Instead of integrating with D4RL \citep{fu2020d4rl} which has dependencies on older versions of MuJoCo, we created our own offline datasets following the protocol outlined in the D4RL paper. For each environment, we created a small offline dataset of 100k transitions by training an RL agent with soft-actor critic as the policy learner and a high entropy coefficient (to encourage exploration). Our datasets  comprise of all transitions in the agent's replay buffer, akin to how the ``medium'' dataset was generated in the D4RL paper. For training, we used the same hyperparameters as suggested in the IQL paper, changing the critic learning rate to $1 \times 10^{-4}$ and the soft update rate of the critic network to 0.05. In Table\ \ref{table:offline rl results} below, we report the normalized scores. We don't aim to match the results of the IQL paper, since we created our own datasets. Still, the results in Table \ref{table:offline rl results} serve as a sanity check and confirm that the IQL implementation in \pearl\, learns a reasonable policy as compared to baseline random policy and the expert policy.


\begin{table}[ht]
  \centering
  \begin{tabular}{|c|c|c|c|c|}
    \hline
    Environment & Random return & IQL return & Expert return & Normalized score \\ \hline
    HalfCheetah-v4 & -426.93 & 145.89 & 484.80 & 0.62 \\
    Walker2d-v4 & -3.88 & 1225.12 & 2348.07 & 0.52 \\
    Hopper-v4 & 109.33 & 1042.03 & 3113.23 & 0.31 \\
  \hline
\end{tabular}
  \caption{Normalized scores of Implicit Q-learning on three different continuous control MuJoCo environments. Here, ``Random return'' refers to the average return of an untrained SAC agent, ``IQL return'' refers to the average evaluation returns of the trained IQL agent (episodic returns of the trained agent when interacting with the environment) and ``Expert return'' is the maximum episodic return in the offline dataset. The normalized score is computed as the ratio of IQL return and Expert return, taking Random return as the baseline, as proposed in \cite{kostrikov2021offline}.}
  \label{table:offline rl results}
\end{table}


\subsection{Neural Contextual Bandits (CB) Benchmarks}
\label{subsec:cb_benchmarks}
\begin{figure}[!htb]
\minipage{0.50\textwidth}
    \includegraphics[width=\linewidth]{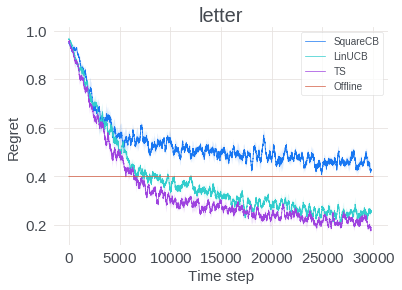}
\endminipage\hfill
\minipage{0.50\textwidth}
    \includegraphics[width=\linewidth]{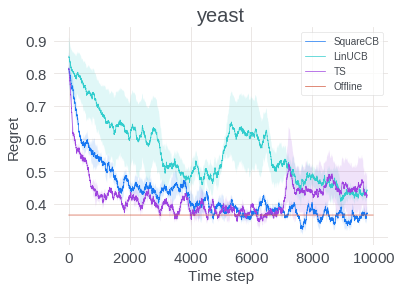}
\endminipage\hfill
\minipage{0.50\textwidth}%
    \includegraphics[width=\linewidth]{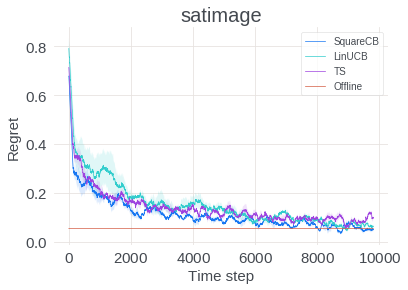}
\endminipage
\minipage{0.50\textwidth}%
    \includegraphics[width=\linewidth]{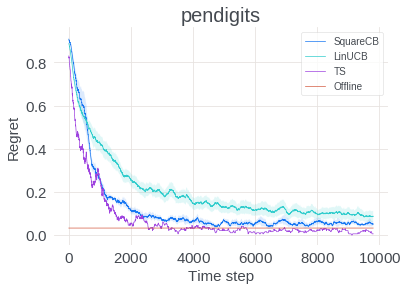}
\endminipage
    \caption{Performance of neural contextual bandit policy learner with Linear UCB, Thompson Sampling and SquareCB based exploration modules in \pearl\, on four UCI datasets. We also add an offline baseline that is considered near optimal.}
\label{fig:cb_benchmarks}
\end{figure}



\looseness-1 We implemented and tested neural adaptations of common CB algorithms, including LinUCB \citep{li2010contextual}, Thompson Sampling (TS) \citep{agrawal2013thompson}, and SquareCB~\citep{foster2020beyond}. 
We give a few details of our implementations. Let $(x,a)$ denote a (context, action) pair. The neural adaptions of LinUCB and TS are based on calculating an exploration bonus term $\| \phi_t(x,a)\|_{A_t^{-1/2}}$ where $\phi_t(x,a)$ is the feature representation\footnote{We took the feature representation to be the output of the last layer of a neural network.} of $(x,a)$ at learning step $t$ and $A_t= I + \sum_{n=1}^{t-1} \phi_t(x,a)\phi_t(x,a)^T$ is the feature covariance matrix. 

\vspace{4pt}
For LinUCB, we explicitly added the term $\beta \times \| \phi_t(x,a)\|_{A_t^{-1/2}}$ as the exploration bonus\footnote{We set $\beta=0.25$. This choice empirically improved performance.} to the predicted reward for each $(x,a)$. For the neural TS, for each $(x,a)$, we sampled reward from a Gaussian distribution, with mean equal to the reward model prediction and variance equal to $\|\phi_t(x,a)\|_{A_t^{-1/2}}$.
For the SquareCB method, we followed the same action selection rule as described in \cite{foster2020beyond}.
We set the scaling parameter as $\gamma = 10 \sqrt{dT}$ where $d$ is the dimension of the feature-action input vector, $\phi(x,a)$. For all three algorithms, actions were represented using binary encoding. Hence, dimension of the feature-action input vector is $d_x + \log_2(A)$, where $d_x$ is the dimension of the context feature $\phi(x)$ and $A$ is the number of actions\footnote{We consider a model where feature-action vector $\phi(x,a)$ is a concatenation of the context feature vector $\phi(x)$ and the action representation.}.
Note that \cite{foster2020beyond} recommend setting $\gamma \propto \sqrt{dT}$ for the linear bandit problem. We scaled this by 10 as we empirically observed improved performance in our ablation study.



To design our experiments, we followed a common way of using supervised learning datasets to construct offline datasets for CB settings \citep{dudik2011doubly,foster2018practical,bietti2021contextual}. We used the UCI repository \citep{UCIMLRepository} to select the supervised learning datasets. Each of these datasets is a collection of features $x$ and the corresponding true labels $y$. We make the following transformation to construct the offline CB datasets. Given a pair $(x,y)$ drawn from a dataset, we set the reward function as $r(x,a)= \mathbbm{1}\{ a = y \}+\epsilon$, where $\epsilon \sim \mathcal{N}(0,\sigma)$. That is, the agent receives an expected reward of 1 when it chooses the correct label $a=y$, and gets an expected reward of 0 otherwise. 

Besides testing different contextual bandit algorithms, we implemented an offline learner as a baseline (see the ``Offline'' label in Figure\ \ref{fig:cb_benchmarks}). The offline learner only has access to a diverse offline dataset, using which it estimates a reward model $\hat{r}(x,a)$. In  Figure~\ref{fig:cb_benchmarks}, we show the performance of a greedy policy with respect to the reward model, i.e. $\pi(x)\in \arg\max_a \hat{r}(x,a)$.


The plots presented in Figure\ \ref{fig:cb_benchmarks} show the average learning curves across 5 runs for each of the four datasets, \textit{letter}, \textit{yeast}, \textit{satimage} and \textit{pendigits}. The confidence intervals represent one standard error. These curves show that all the tested algorithms (UCB, TS and Square CB) matched the performance of the baseline.\footnote{For the \textit{letter} dataset LinUCB and TS surpassed the baseline's performance.} This validates our implementation of these intelligent exploration methods for policy learning in the contextual bandit problems.




\paragraph{Experiment Setup for Contextual Bandit Algorithms.} For the \textit{letter}, \textit{satimage} and \textit{pendigits} datasets, our experiments used an MLP with two hidden layers of size $64 \times 16$. For the \textit{yeast} dataset, our experiments used an MLP with two hidden layers of size $32 \times 16$. We used the Adam optimizer with a learning rate of $0.01$ and used a batch size of 128. 
For action representation, we used a binary decoding of the action space, i.e. we represented each action as its binary representation required to represent all actions.



\subsection{Versatility Benchmarks}
\label{subsec:versatility_benchmarks}
This section provides details of some benchmarking experiments we ran to test different features in \pearl, namely a) to learn in partially observable settings by summarizing history, b) to effectively and efficiently explore in sparse reward settings, c) to learn risk averse policies, d) to learn with constraints on long-run costs, and e) to learn in problem settings with dynamic action spaces. We do not perform hyperparameter search to optimize performance and the experiments are meant to only demonstrate that \pearl\, can be used effectively to learn a good policy in different challenging scenarios. 


\paragraph{History summarization for problems with partially observable states.}

To test the history summarization module\, in \pearl, we adapted CartPole, a classic control environment in OpenAI Gym, to design a variant where the state is partially observable. The goal is still the same --- to keep a pole upward by moving a cart linked to the pole. In the original CartPole environment, the agent can perceive the true underlying state --- angle and angular velocity of the pole, as well as the position and velocity of the cart. In the environment we design, only the angle of the pole and position of the cart are observable. Consequently, an RL agent must use current and past observations to deduce the velocities, since both angular velocity of the pole and velocity of the cart are crucial for selecting the optimal action. To further increase the degree of partial observability, the new environment was designed to emit the observation (angle of the pole and position of the cart) after every 2 interactions, instead of every interaction. For other interactions, the environment emits a vector of zeros.

\begin{figure*}[h!]
\vspace{-8pt}
    \centering
    \begin{subfigure}[t]{0.33\textwidth}
        \includegraphics[width=1.1\textwidth]{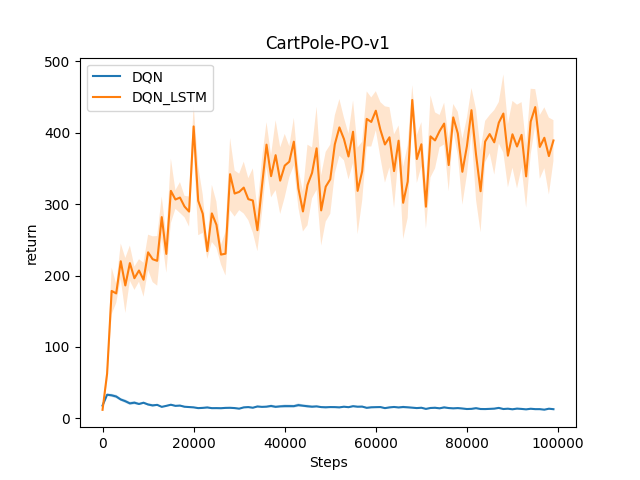}
        \subcaption{}
        \label{fig:classic control partial observable results}
    \end{subfigure}\hfill%
    \begin{subfigure}[t]{0.33\textwidth}
        \includegraphics[width=1.1\textwidth]{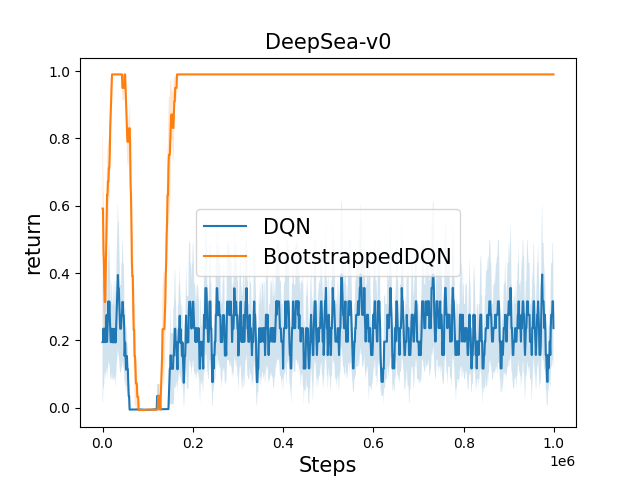}
        \subcaption{}
        \label{fig:classic control sparse reward result}
    \end{subfigure}
    \begin{subfigure}[t]{0.33\textwidth}
        \includegraphics[width=1.1\textwidth]{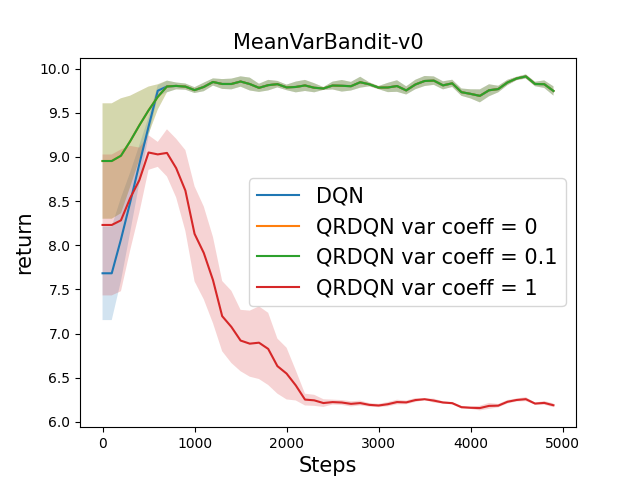}
        \subcaption{}
        \label{fig:classic control safety results}
    \end{subfigure}\hfill%
    \vspace{-5pt}
    \caption{Agent Versatility Benchmark Results: learning curves for (a) DQN, with and without LSTM, in the partially observable CartPole environment, (b) DQN and Bootstrapped DQN in a $10\times10$ Deep Sea environment, and (c) QR-DQN with mean variance risk sensitive safety module. Under a high degree of risk aversion, the learned policy has a smaller expected return but a smaller variance of the return distribution as well.}
    \vspace{-6pt}
\end{figure*}

We benchmarked two RL agents on this new environment---one with and other without the LSTM history summarization module. Both used DQN as the policy learner along with an $\epsilon$-greedy exploration module with $\epsilon=0.1$. Other hyperparameters were taken to be the same as described in Section \ref{subsec:RL_benchmarks}. We used an LSTM with 2 recurrent layers, each with 128 hidden units (i.e. dimension of the hidden state is taken to be 128). The history length was chosen to be $4$. Figure\ \ref{fig:classic control partial observable results} plots the mean and standard deviation of episodic returns over five runs. As can be clearly seen, the agent which used the LSTM-based history summarization module learned to achieve high returns\footnote{The maximum episodic return in the original CartPole environment and the new CartPole environment we design is 500. } quickly while the other agent failed to learn. This simple example illustrates how an RL agent in \pearl\, can be equipped with the ability to use historical information to learn in a partially observable environment.

\vspace{-5pt}
\paragraph{Effective exploration for sparse reward problems.}


In section \ref{subsec:cb_benchmarks}, we benchmarked intelligent exploration algorithms in \pearl\, used for bandit learning. Here we tested the Bootstrapped DQN algorithm \citep{osband2016deep} implemented in \pearl, which enables intelligent exploration in RL settings. To do so, we used the \emph{DeepSea} environment \citep{osband2019deep}. DeepSea is a sparse reward environment and is well known to be challenging for exploration. It has $n \times n$ states and is fully deterministic. For our experiments, we chose $n = 10$, for which the chance of reaching the target state under a random policy is $2^{-10}$. 

We tested two RL agents -- one with a DQN policy learner and $\epsilon$-greedy exploration ($\epsilon=0.1$) and the other with Bootstapped DQN. The hyperparameters for DQN and Bootstrapped DQN were taken to be the same as in Section\ \ref{subsec:RL_benchmarks}. For each experiment, we performed 5 runs, each with $10^6$ steps. Figure \ref{fig:classic control sparse reward result} shows the learning curves of the two agents. It can be seen that the Bootstrapped DQN quickly learned an optimal policy while the other agent (DQN with $\epsilon$-greedy exploration) failed to learn. Integration of intelligent exploration algorithms in \pearl\, can really help users test out RL agents in real-world problem settings with exploration challenges.

\paragraph{Risk-sensitive policy learning.}
To test out \pearl's ability to learn a policy under different measures of risk sensitivity, we designed a simple environment called \emph{Mean-Variance Bandit}. This environment has only one state and two actions. The reward of each of the two actions follows a Gaussian distribution. The reward distribution for action $1$ has a mean of $6$ and a variance of $1$, while for action $2$, the mean is $10$ and the variance is $9$. Under the classic reward maximization objective, where the goal  is to maximize the expected return, it is easy to see that the optimal policy for this environment will always choose action 2. With risk-sensitive learning however, the system designer can choose to learn a policy by optimizing for a different objective (different than maximizing expected return). 

Risk-sensitive RL works only with distributional learners, like Quantile regression DQN (QR-DQN) \citep{dabney2018distributional} or implicit quantile networks (IQN) \citep{dabney2018implicit}. The basic idea is to learn a distribution over the cumulative reward for each state action pair (denote it by $Z(s,a)$), instead of learning the expectation of them, $Q(s,a)$. Formally, for a given policy $\pi$, $Z$ and $Q$ are defined as follows.
\[
Z^{\pi}(s,a) = \left[ \sum_{t=0}^{\infty} \gamma^t r(s_t, a_t) \mid s_0 = s, a_0 = a, a_t \sim \pi(\cdot \mid s_t) \right], \quad Q^{\pi}(s,a) = \E\left[ Z^{\pi}(s,a) \right].
\]
Note here that $Z^{\pi}(s,a)$ is a random variable. Then, one can use a risk-measure function \citep{ma2020dsac, dabney2018implicit} to map this distribution to Q values. In \pearl, we implemented a risk-measure function that computes Q values ($Q^{\pi}(s,a) = \E \left[ Z^{\pi}(s,a) \right]$); and a mean-variance risk-measure function which computes an weighted average of the mean and the variance of $Z^\pi$, $\E \left[ Z^{\pi}(s,a) \right] - \beta \cdot \text{Var}(Z^{\pi}(s,a))$, where $\text{Var}(Z^{\pi}(s,a))$ denotes the variance of the return distribution and $\beta$ is a parameter that specifies the degree of risk sensitivity\footnote{A value of $\beta=0$ implies the risk neutral measure function which corresponds to the classic expected reward maximization objective in RL. Higher values of $\beta$ imply greater risk aversion.}. 

For the \textit{Mean-Variance Bandit} environment, we tested the ${\tt mean\char`_variance\char`_safety\char`_module}$ in \pearl\,, which uses the mean-variance risk-measure function to compute Q values as shown above. For a non-zero value of $\beta$, a good policy would maximize expected return while minimizing visitations to state action pairs with a high variance of return distribution; and $\beta$ specifies the precise trade-off between these two objectives. As can be easily seen, the optimal policy in the \textit{Mean-Variance Bandit} changes with the value of $\beta$. For any $\beta > 0.5$, the optimal policy always chooses action 1 (the action with lower mean and variance of return) while for $\beta < 0.5$, always choosing action 2 (the action with higher mean and variance of return) is optimal.

For this experiment, we used \pearl's implementation of the QR-DQN algorithm to learn the distribution over the cumulative reward, and attached the ${\tt mean\char`_variance\char`_safety\char`_module}$ as the safety module to learn a risk-sensitive policy. The quantile network used was parameterized by 10 quantiles and  two hidden layers of size 64. We used the AdamW optimizer with a learning rate of $5 \times 10^{-4}$, a batch size of 32, and a replay buffer of size 50,000. We tested three choices of $\beta$, $\beta \in \{0, 0.1, 1\}$). Figure \ref{fig:classic control safety results} shows learning curves for 5000 training steps with different values of $\beta$, averaged over 5 runs. We also show the learning curve for simple DQN\footnote{QR-DQN with $\beta=0$ for the ${\tt mean\char`_variance\char`_safety\char`_module}$ and DQN are different even though both use expected long-run return as Q-values. This is because QR-DQN computes Q-values by first learning the distribution over random returns and then taking an expectation, while DQN directly learns the Q-values.}. As can be clearly seen, for a high value of $\beta$ (which corresponds to a high degree of risk aversion), the learned optimal policy selects action 1, with a mean return of 6. DQN and QR-DQN with smaller values of $\beta$ converged to the optimal policy, which selects action 2, with the mean reward of 10.

While the environmental setup here is simple, it serves as a sanity check for the implementation of the {\tt risk\char`_sensitive\char`_safety\char`_module}\, and distributional policy learner in \pearl. It also illustrates how users can use different risk measures to specify the degree of risk aversion during policy learning. Currently, we have only implemented the mean-variance risk measure function and the QR-DQN algorithm in \pearl. We plan to add more in the next release, for example, implementations of IQN \citep{dabney2018implicit} and Distributional SAC \citep{ma2020dsac}, along with different measures of distorted expectation \citep{ma2020dsac}.

\paragraph{Learning with cost constraints.}

\begin{figure}[t]
\minipage{0.25\textwidth}
    \includegraphics[width=\linewidth]{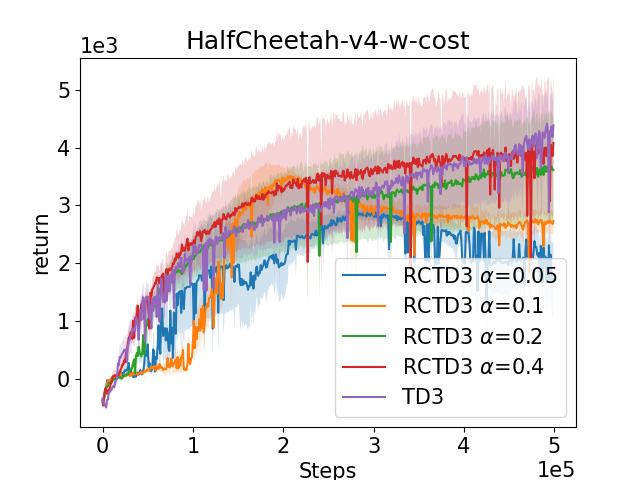}
\endminipage\hfill
\minipage{0.25\textwidth}
    \includegraphics[width=\linewidth]{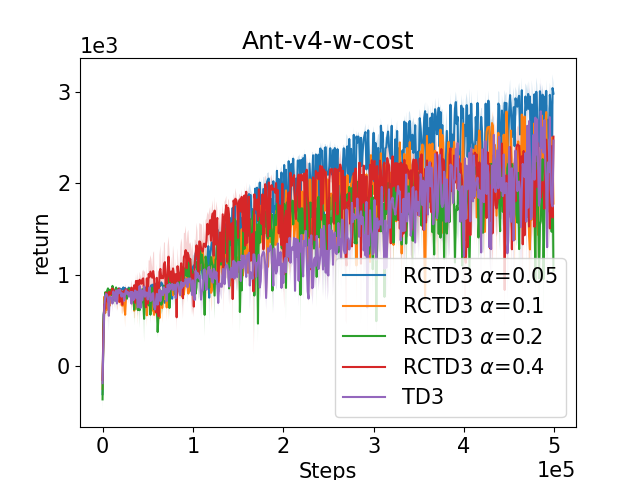}
\endminipage\hfill
\minipage{0.25\textwidth}%
    \includegraphics[width=\linewidth]{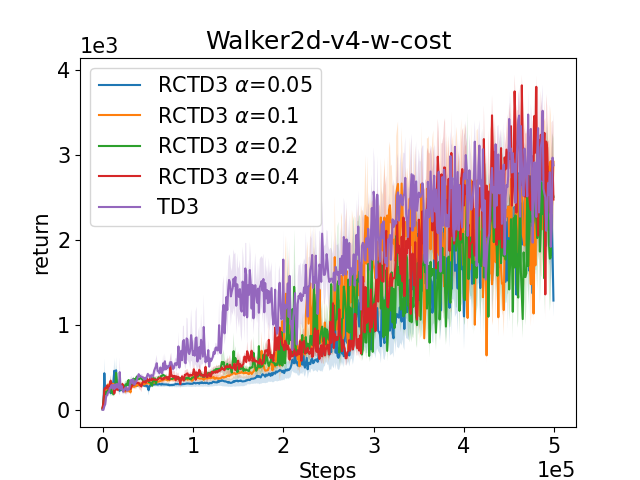}
\endminipage
\minipage{0.25\textwidth}%
    \includegraphics[width=\linewidth]{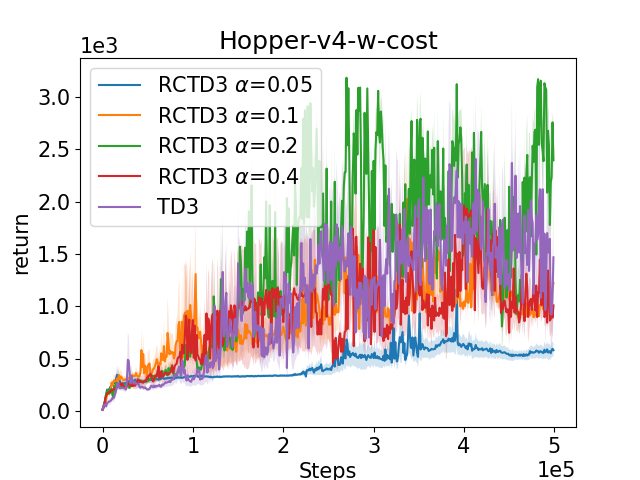}
\endminipage

\minipage{0.25\textwidth}
    \includegraphics[width=\linewidth]{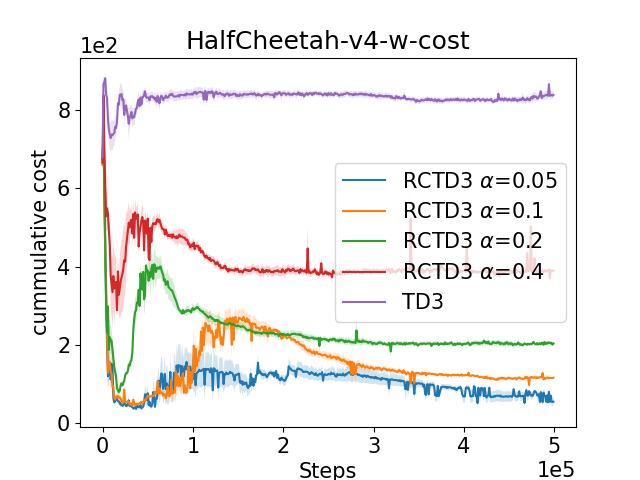}
\endminipage\hfill
\minipage{0.25\textwidth}
    \includegraphics[width=\linewidth]{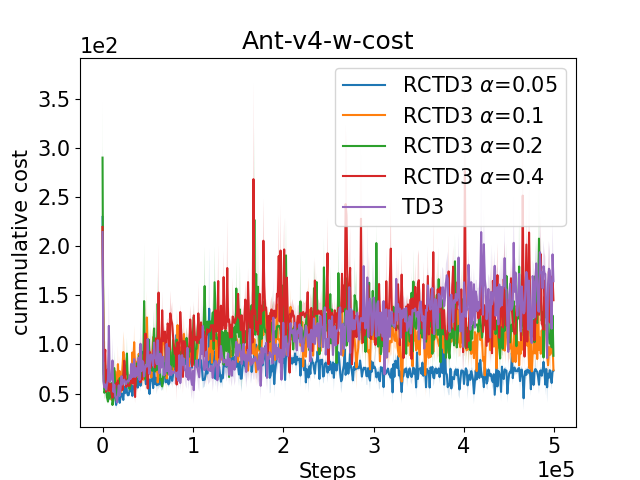}
\endminipage\hfill
\minipage{0.25\textwidth}%
    \includegraphics[width=\linewidth]{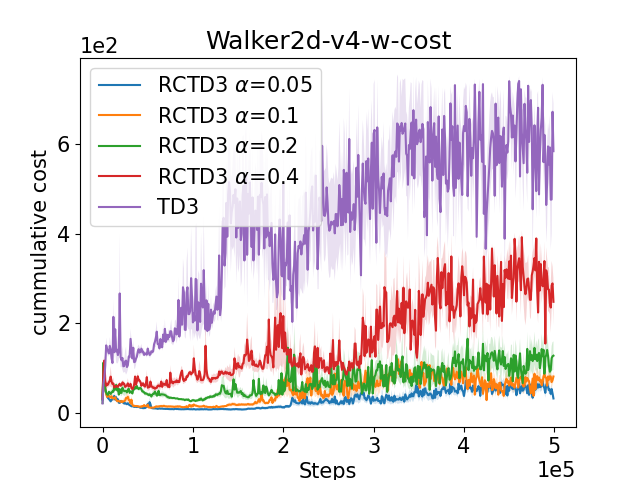}
\endminipage
\minipage{0.25\textwidth}%
    \includegraphics[width=\linewidth]{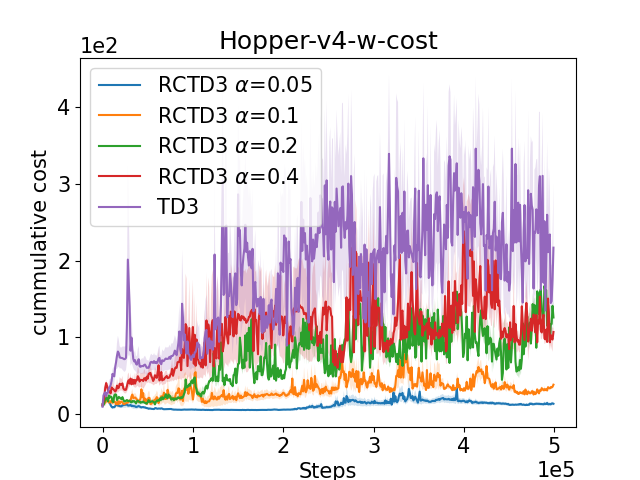}
\endminipage

\minipage{0.25\textwidth}
    \includegraphics[width=\linewidth]{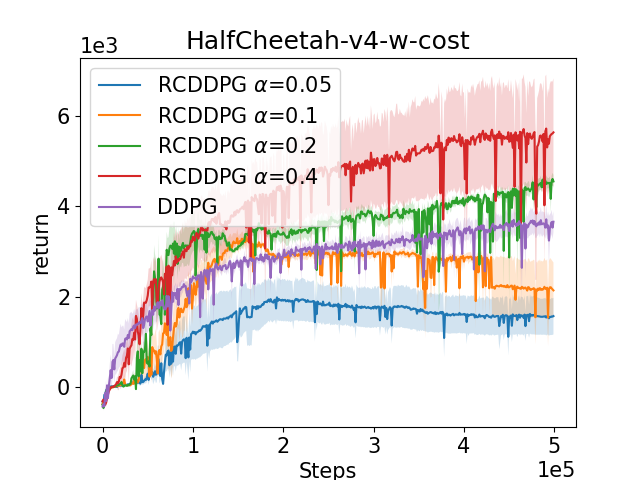}
\endminipage\hfill
\minipage{0.25\textwidth}
    \includegraphics[width=\linewidth]{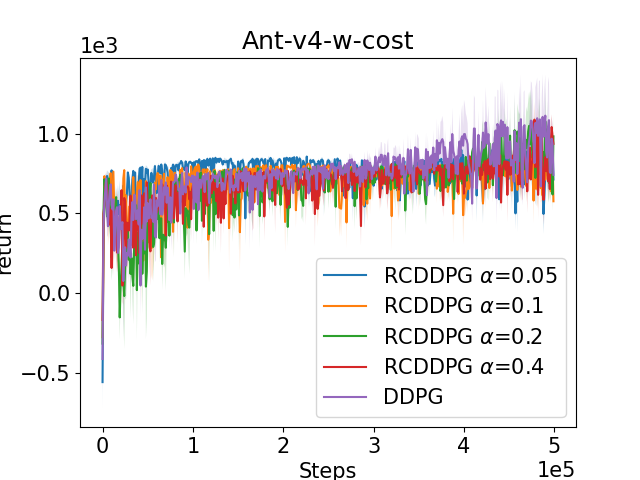}
\endminipage\hfill
\minipage{0.25\textwidth}%
    \includegraphics[width=\linewidth]{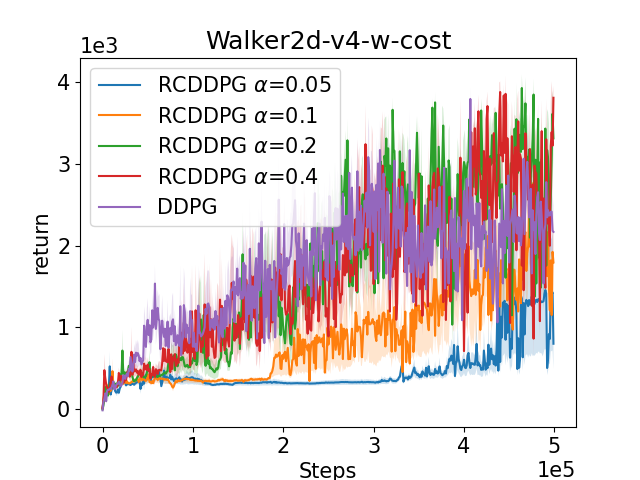}
\endminipage
\minipage{0.25\textwidth}%
    \includegraphics[width=\linewidth]{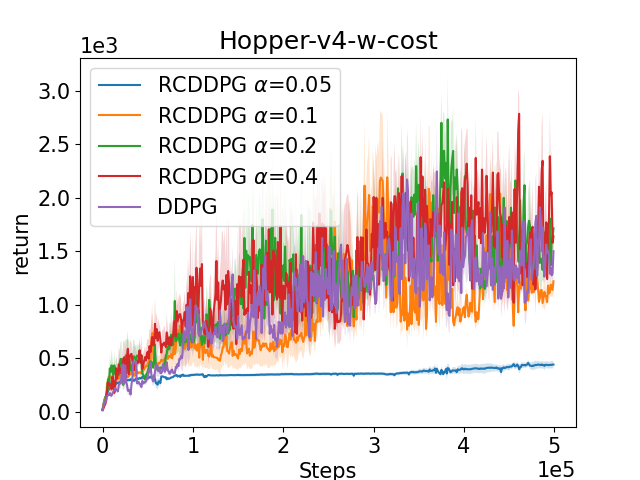}
\endminipage

\minipage{0.25\textwidth}
    \includegraphics[width=\linewidth]{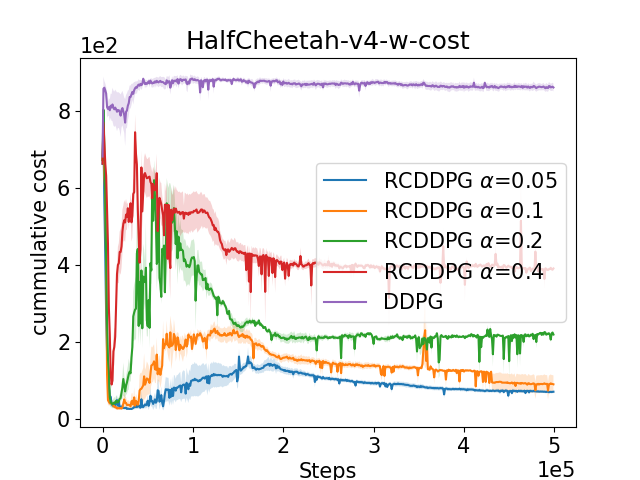}
\endminipage\hfill
\minipage{0.25\textwidth}
    \includegraphics[width=\linewidth]{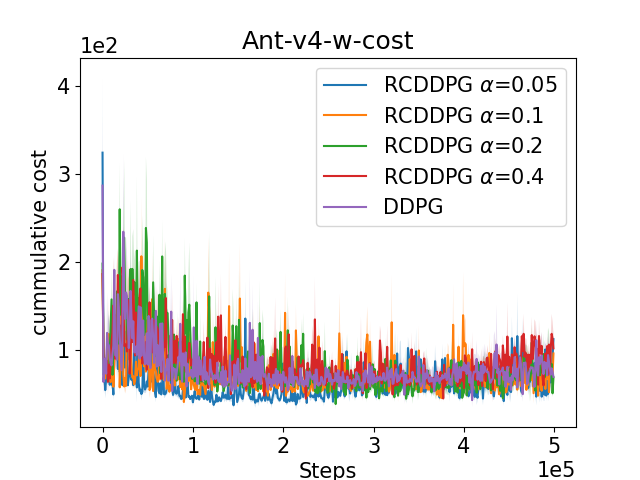}
\endminipage\hfill
\minipage{0.25\textwidth}%
    \includegraphics[width=\linewidth]{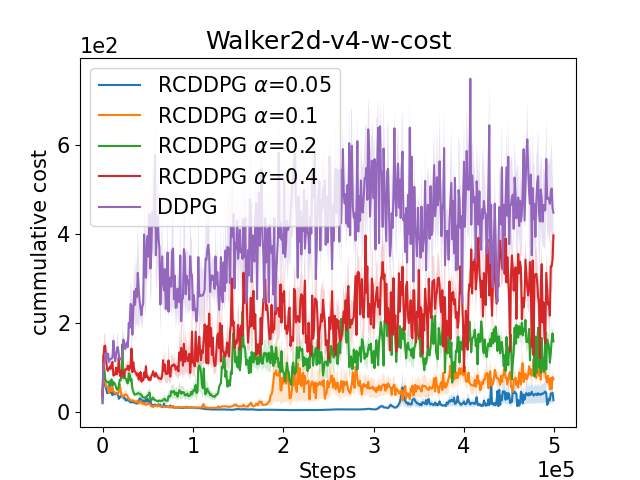}
\endminipage
\minipage{0.25\textwidth}%
    \includegraphics[width=\linewidth]{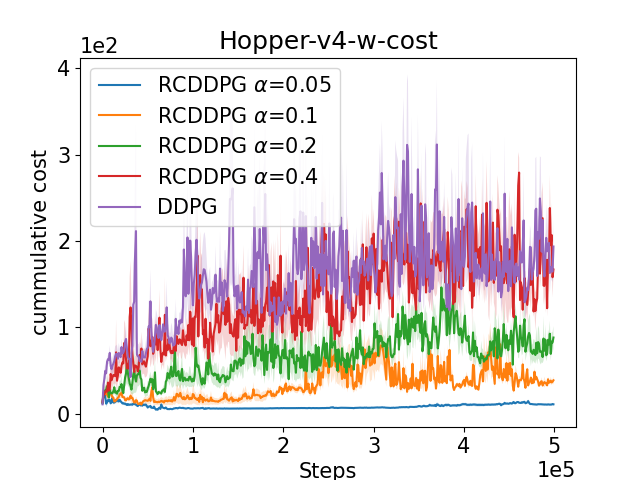}
\endminipage

\minipage{0.25\textwidth}
    \includegraphics[width=\linewidth]{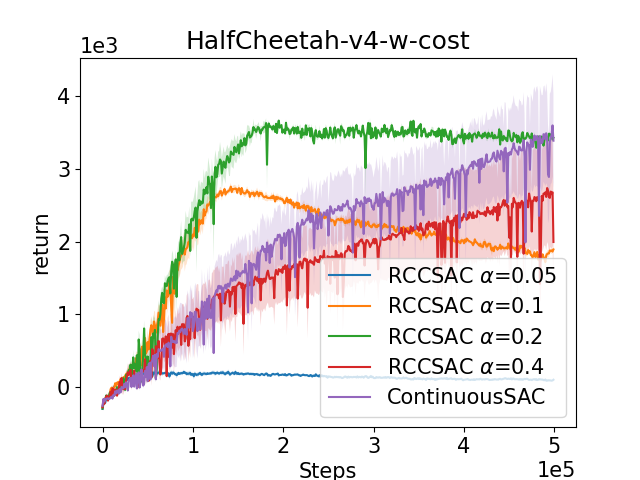}
\endminipage\hfill
\minipage{0.25\textwidth}
    \includegraphics[width=\linewidth]{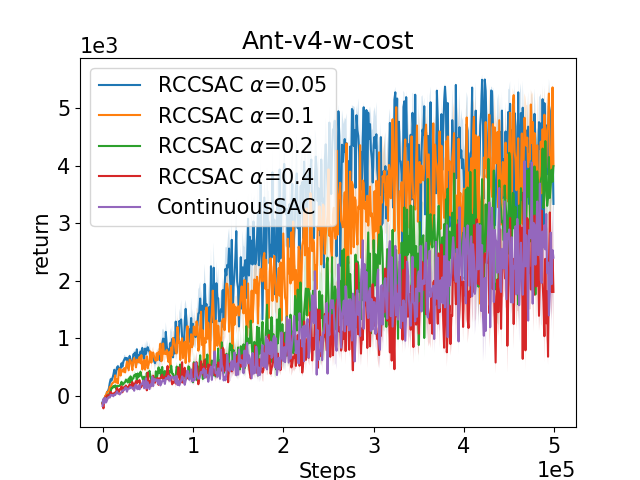}
\endminipage\hfill
\minipage{0.25\textwidth}%
    \includegraphics[width=\linewidth]{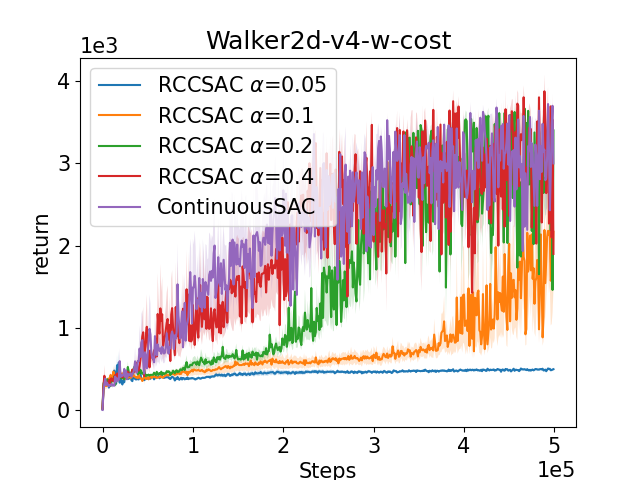}
\endminipage
\minipage{0.25\textwidth}%
    \includegraphics[width=\linewidth]{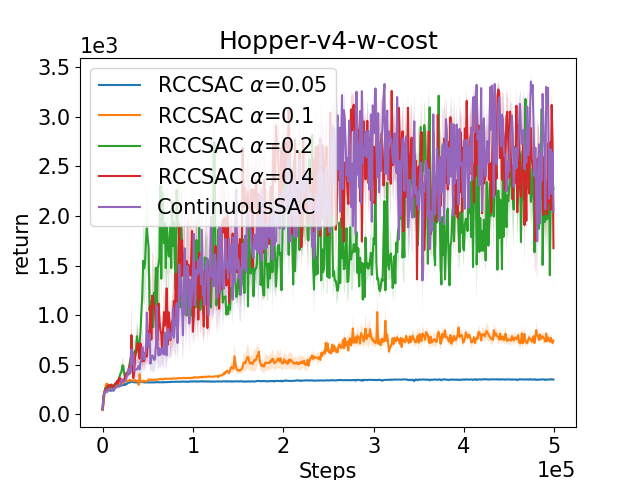}
\endminipage

\minipage{0.25\textwidth}
    \includegraphics[width=\linewidth]{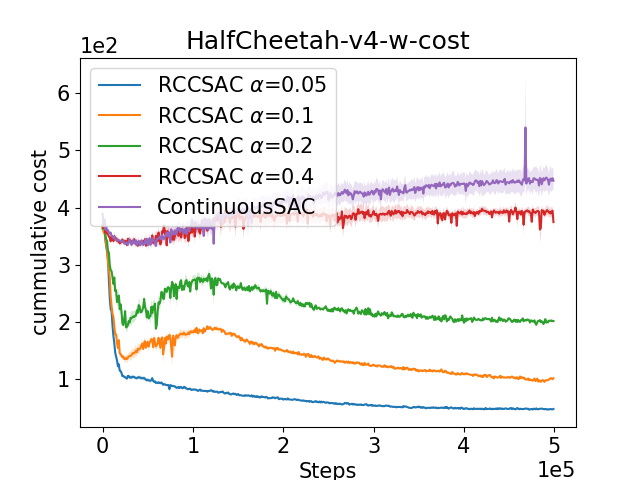}
\endminipage\hfill
\minipage{0.25\textwidth}
    \includegraphics[width=\linewidth]{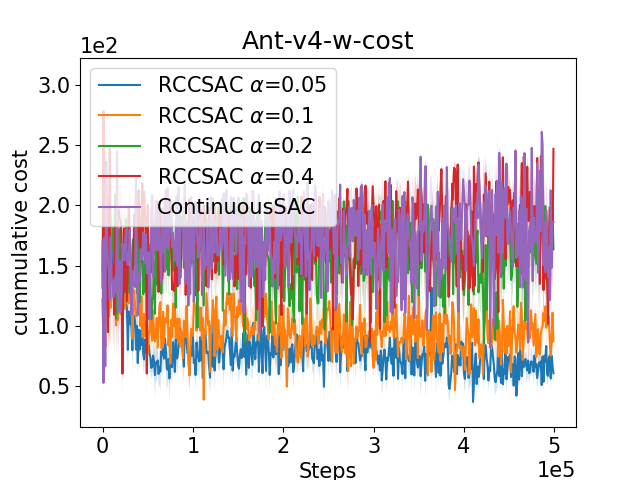}
\endminipage\hfill
\minipage{0.25\textwidth}%
    \includegraphics[width=\linewidth]{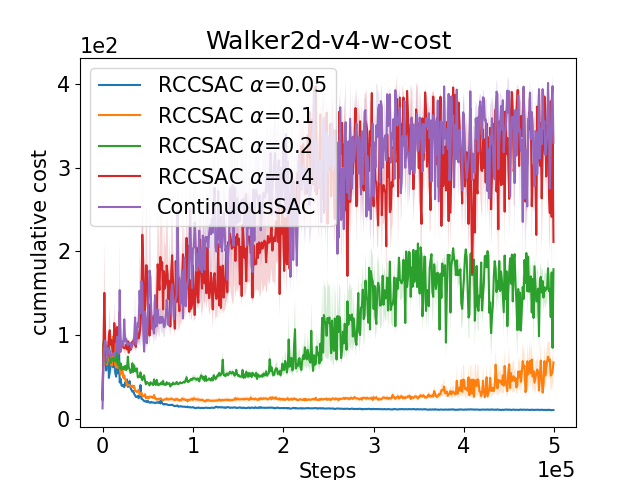}
\endminipage
\minipage{0.25\textwidth}%
    \includegraphics[width=\linewidth]{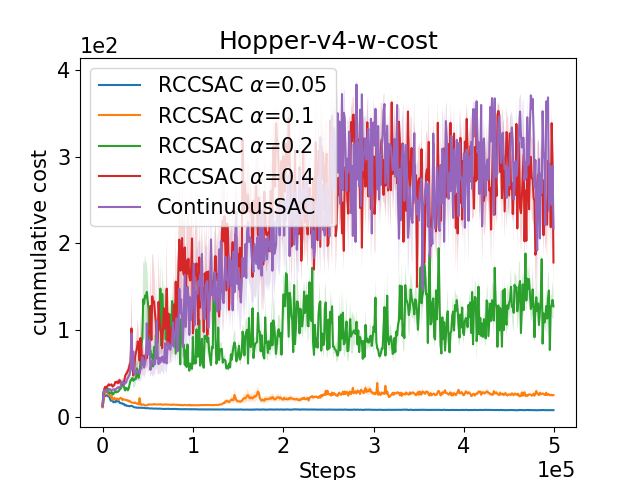}
\endminipage

\caption{Episodic cost (top) and episodic return (bottom) plots during training on continuous control tasks with cost and reward feedback. The plots present performance of TD3, DDPG, and CSAC and our cost constraint adaptations of RCTD3, RCDDPG, and RCCSAC for multiple values of constraint threshold $\alpha$. See text for details.}
\label{fig:rcpo}
\end{figure}


Many real-world problems require an agent to learn an optimal policy subject to cost constraint(s). Moreover, for problems where a reward signal is not well defined, it is often useful to specify desirable behaviour in the form of constraints. For example, limiting the power consumption of a motor can be a desirable constraint for learning robotic locomotion. 

Optimizing for long-run expected reward subject to constraint(s) requires modification to the learning procedure. This setting is often formulated as a constrained MDP. \pearl\, enables learning in constrained MDPs through the {\tt reward\char`_constrained\char`_safety\char`_module} which implements the Reward Constrained Policy Optimization algorithm of \cite{tessler2018reward}.




To test our implementation, we modified a gym environment with a per-step cost function, $c(s, a)$, in addition to the standard per-step reward $r(s, a)$. We chose the per-step cost to be $c(s,a) = || a ||_2^2$ where $a\in \mathbb{R}^d$, which approximates the energy spent in taking action $a$ in state $s$.
Figure~\ref{fig:rcpo} shows the results of different actor-critic algorithms with a constraint on the (normalized) expected discounted cumulative costs. Specifically, for different values of the threshold $\alpha$, we add a constraint
\[
(1-\gamma)\,\E_{s_t \sim \eta_{\pi}, a_t \sim \pi} \left[ \sum_{t=0}^\infty \gamma^t c(s_t, a_t) \, | \, s_0 = s, a_0 = a \right] \leq \alpha.
\]
We experimented with reward constrained adaptations of different actor-critic policy learners like TD3, DDPG and Continuous SAC (CSAC), by instantiating corresponding RL agents in \pearl\, and specifying the safety module to be the {\tt reward\char`_constrained\char`_safety\char`_module}. These are referred to as reward constrained TD3 (RCTD3), RCDDPG and RCCSAC respectively. Hyperparameter values for the policy learners were chosen to be the same as specified above in subsection \ref{subsec:RL_benchmarks}. The {\tt reward\char`_constrained\char`_safety\char`_module} has only one hyperparameter, $\alpha$. We experimented with different values of $\alpha$ as shown in Figure \ref{fig:rcpo}.
 
Figure \ref{fig:rcpo} shows episodic cumulative rewards and costs for four continuous control MuJoCo tasks. As can be seen, the cumulative costs are smaller for a smaller value of the threshold $\alpha$. Interestingly, moderate values of $\alpha$ in different environments did not result in a significant performance degradation, despite addition of the cost constraint---in fact, for some environments, the cost constraint led to policies with slightly improved episodic returns (for example, RCDDPG in Half-Cheetah).

\paragraph{Adapting to dynamic action spaces.}
\begin{figure*}[h!]
\vspace{-8pt}
    \centering
    \begin{subfigure}[t]{0.5\textwidth}
        \includegraphics[width=\textwidth]{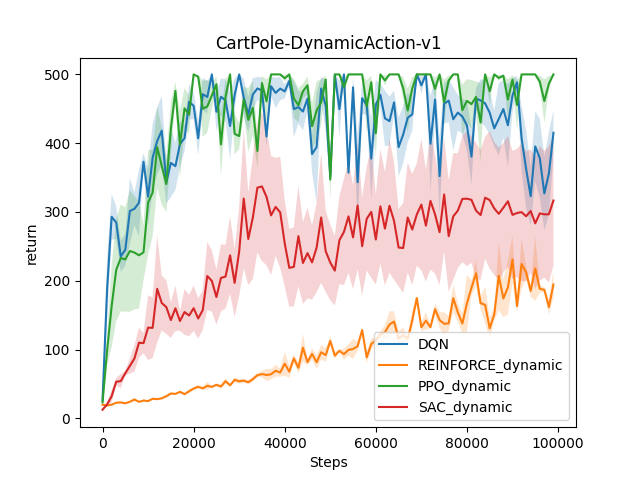}
        \subcaption{}
    \end{subfigure}\hfill%
    \begin{subfigure}[t]{0.5\textwidth}
        \includegraphics[width=\textwidth]{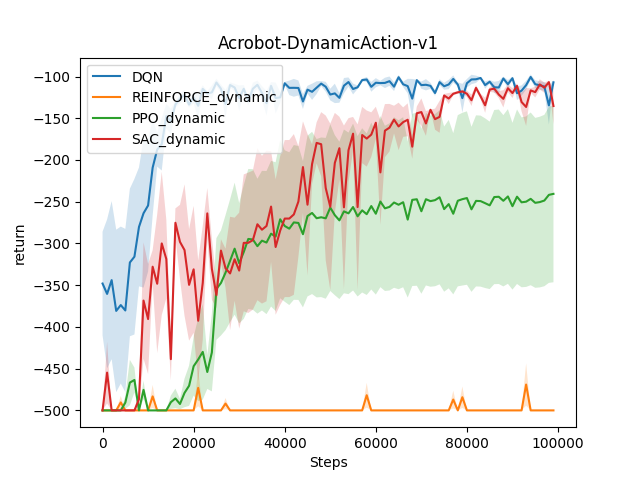}
        \subcaption{}
    \end{subfigure}
    \vspace{-5pt}
    \caption{Benchmark results for dynamic action space: Return of DQN, SAC, PPO and REINFORCE on CartPole and Acrobot environments where each environment deletes an action from the action space, every 4 steps. Note that actor-critic algorithms like SAC, PPO and REINFORCE require a specialized policy network architecture to learn with a dynamic action space.
    }
    \label{fig:dynamic_action_results}
\end{figure*}

In many problems, agents are required to learn in environments where the action space changes at each time step. Practitioners commonly encounter this in recommender systems\footnote{See \cite{chen2022off} for example, and the description of the \textit{Auction based recommender system} and the \textit{Creative selection} projects in Section \ref{sec:product_adoptions}.}, where for every user interaction, the agent encounters a distinct set of content (to choose from) to recommend. To evaluate Pearl's adaptability to dynamic action spaces, we crafted two environments based on CartPole and Acrobot. In these environments, after every four steps, the agent loses access to action 1 in CartPole and action 2 in Acrobot, respectively.

Figure \ref{fig:dynamic_action_results} shows the learning curves for DQN, SAC, PPO, and REINFORCE within these specialized environments. For these experiments, we used the same set of hyperparameters as in Section\ \ref{subsec:RL_benchmarks}. Despite the increased complexity posed by dynamic action spaces, most agents successfully learned effective policies after 100,000 steps, except for REINFORCE consistently under performed in comparison to other algorithms. These simple experiments illustrate that \pearl\, can be used seamlessly to learn in settings with dynamic action spaces as well.

\section{Additional Details Regarding Pearl Architecture}
\subsection{Support for Both Offline and Online Reinforcement Learning}
As mentioned in section \ref{sec: pearl_agent}, Pearl supports both offline and online policy learning algorithms. Offline learning algorithms, for example Conservative Q-learning (CQL) \citep{kumar2020conservative} and Implicit Q-learning (IQL) \citep{kostrikov2021offline}, are designed to learn from offline data and avoid extrapolation errors by adding some form of conservatism when learning the value function. In Pearl, both the offline and online policy learning algorithms are implemented as different policy learners. 

It is noteworthy that Pearl's modular design allows users to combine offline learning and online learning in the training pipeline. A naive way to implement such an offline-to-online RL transition is to train a Pearl agent using offline data and an offline policy learning algorithm (say, CQL) and then use the learned policy and value function (policy network and value function network) to initialize an online Pearl agent. This way, users can effectively leverage offline data to warm start online learning. 

Besides the naive way mentioned above, there is recent literature proposing novel ideas to perform offline to online RL transition, see \citep{yu2023actor, lee2022offline, nakamoto2024cal}. However, this is an active area of research. As part of our ongoing work, we plan to integrate some of these ideas in Pearl. We think this should be easy to do given Pearl’s modular design. 

\subsection{Support for Intelligent Exploration Beyond Dithering}
Our design of the exploration module in Pearl along with its tight integration with the policy learner module does provide support for intelligent exploration strategies beyond dithering. For example, Bootstrapped DQN \citep{osband2016deep} is implemented in this Github {\color{blue}\href{https://github.com/facebookresearch/Pearl/blob/f84334cbff6c698ddca8108530ebfbbc852261aa/pearl/policy_learners/exploration_modules/sequential_decision_making/deep_exploration.py}{file}}. We designed the policy learner and exploration module to work closely with each other - the policy learner module in Pearl computes the necessary quantities, such as value functions, representations and policy distributions. An exploitative action can be taken if the agent is not exploring. For the case when the agent wants to explore, the exploration module can use the quantities computed by the policy learner to determine the exploratory action. For instance, in the case of Bootstrapped DQN, the policy learner learns an approximate posterior distribution over estimates of the optimal value function by bootstrapping and the exploration module implements deep exploration (Thompson sampling) by sampling a from it to take actions in each episode.

\subsection{Support for Dynamic Action Spaces}
We would like to note that Pearl is the first open-source RL library that provides support on dynamic action spaces for both value-based and actor-critic methods. 

Previously, only ReAgent \citep{gauci2018horizon} provided limited support for dynamic action spaces with only value based methods (Deep Q-learning). We note that the implementation of dynamic action spaces in ReAgent has flaws -- available actions in each state are taken into account only at the inference time (i.e. dynamic action space is used only during agent-environment interactions rather than being used both during training and inference). In addition, ReAgent does not support dynamic action spaces with policy gradient or actor-critic methods. Pearl provides support for dynamic action spaces through innovations for both value-based and actor-critic methods.

\subsubsection{Dynamic Action Space Support for Value-Based Learning Methods}\label{sec:value-based-dynamic-action}
\begin{figure}
    \centering
    \includegraphics[width=0.9\textwidth]{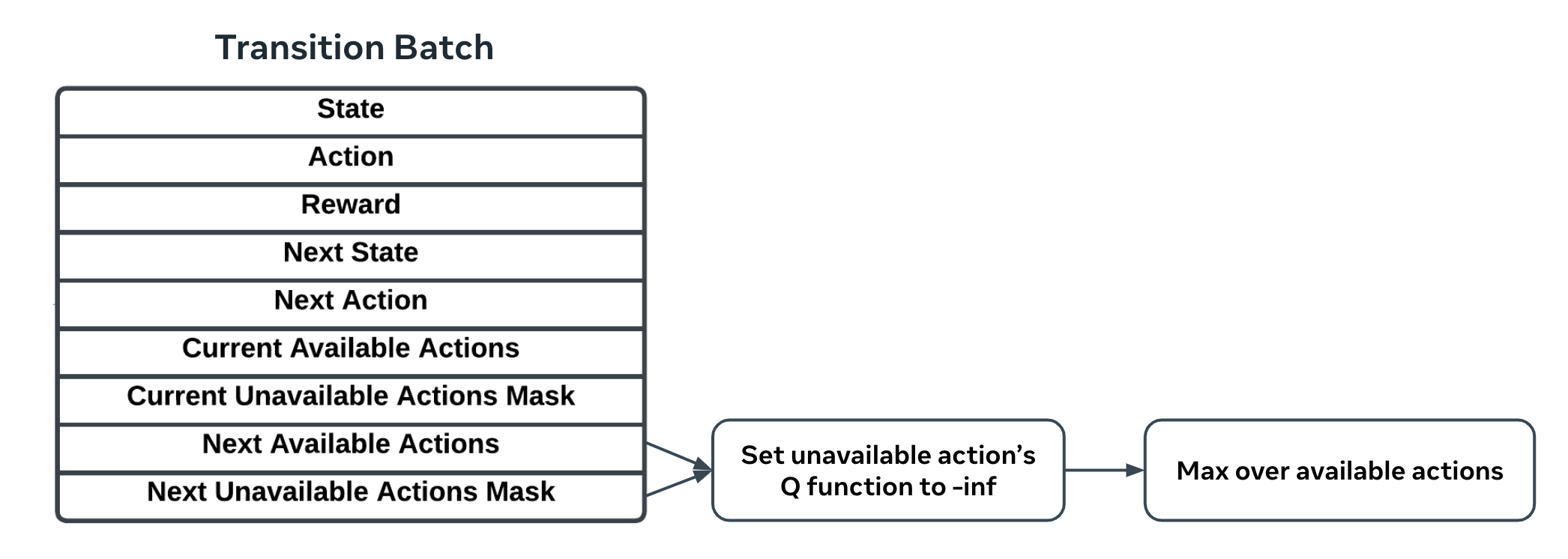}
    \caption{Value-Based RL with Dynamic Action Space Design}
    \label{fig:value-based-dynamic-action}
\end{figure}

The key idea to support dynamic action spaces for value-based methods is to calculate the optimization target in the Bellman equation based on the maximum Q-values over the next state's action space. Mathematically, let $(S_t, A_t, r(S_t, A_t), S_{t+1})$ be a transition tuple. Then, the value function of the optimal policy can be estimated by iteratively minimizing the error between the left and right hand side of the following Bellman equation,
\begin{align*}
    Q_{\theta}(S_t, A_t) = r(S_t, A_t) + \gamma \max_{a \in \mathcal{A}_{S_{t+1}}}Q_{\theta}(S_{t+1}, a),
\end{align*}
where we denote $\mathcal{A}_{S_{t+1}}$ as the action space associated with state $S_{t+1}$. To achieve this, we designed a replay buffer to store two additional attributes per state, the {\tt available\_actions} and the {\tt available\_actions\_mask}. Please refer to Figure \ref{fig:value-based-dynamic-action} which shows addition of {\tt available\_actions} and {\tt available\_actions\_mask} for the current and next state in a transition tuple. 

The {\tt available\_actions} attribute is a tensor of representations of available actions. Here, we use zero padding for unavailable actions. The {\tt available\_actions\_mask} is simply a tensor with a mask for unavailable actions. To calculate the optimization targets (right hand side of the Bellman equation), $\max_{a \in \mathcal{A}_{S_{t+1}}}Q_{\theta}(S_{t+1}, a)$, we first calculate $Q_{\theta}(S_{t+1}, a)$ for all actions $a$ (including unavailable actions using the zero padded representations). We then use the {\tt available\_actions\_mask} to set Q-values of unavailable actions to $-\infty$ to ensure that these do not contribute to gradient updates when minimizing the Bellman error.

\subsubsection{Dynamic (Discrete) Action Space Support for Actor-Critic Methods}
For policy gradient and actor-critic methods, Pearl only supports dynamic discrete action spaces, following the work of \cite{chen2022off}. There seems to be no literature on policy optimization with continuous action spaces that change with different states.

\begin{figure}
    \centering
    \includegraphics[width=0.9\textwidth]{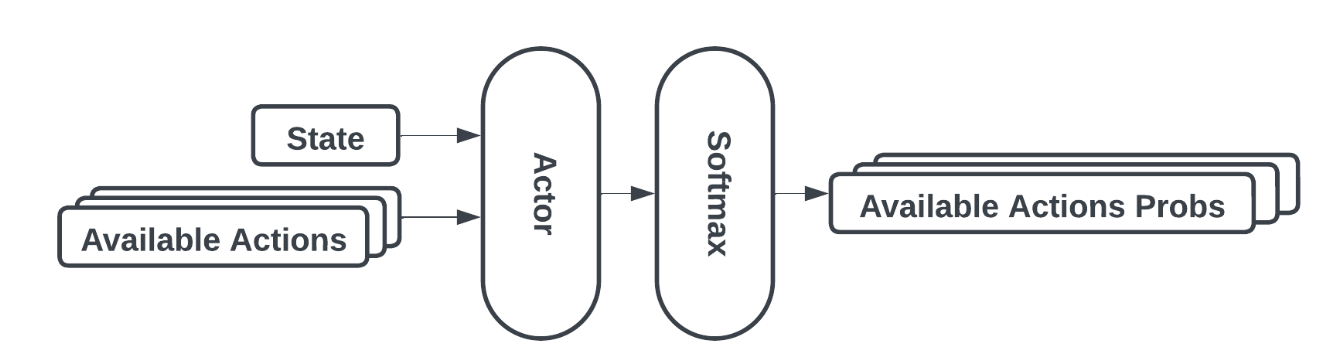}
    \caption{Dynamic Actor Network for Dynamic Action Policies}
    \label{fig:dynamic-actor-network}
\end{figure}

To enable policy learning using actor-critic methods in problems with dynamic action spaces, we implement a policy neural network architecture called dynamic actor network, as illustrated in Figure \ref{fig:dynamic-actor-network}, where the action probabilities are computed as 
\begin{align*}
    \pi_\phi(a|s) = \frac{f_\phi(s, a)}{\sum_{a' \in \mathcal{A}_s}f_\phi(s, a')}.
\end{align*}
Here, we denote $\pi_\phi$ as the dynamic actor network which outputs action probabilities for all actions in the action space $\mathcal{A}_s$ (state dependent action space) and let $f_\phi$ denote the penultimate layer of the actor network. The action probabilities are computed by applying softmax to the final layer, $f_\phi$. 
Given this, we can write the Bellman equation for critic learning as
\begin{align*}
    Q_\theta(S_t, A_t) = r(S_t, A_t) + \gamma\mathbb{E}_{a \sim \pi_\phi}[Q_\theta(S_{t+1}, a)],
\end{align*}
and can use the learned critic, $Q_\theta$, to update the policy network using policy gradient or actor-critic algorithms. 

\begin{figure}
    \centering
    \includegraphics[width=0.9\textwidth]{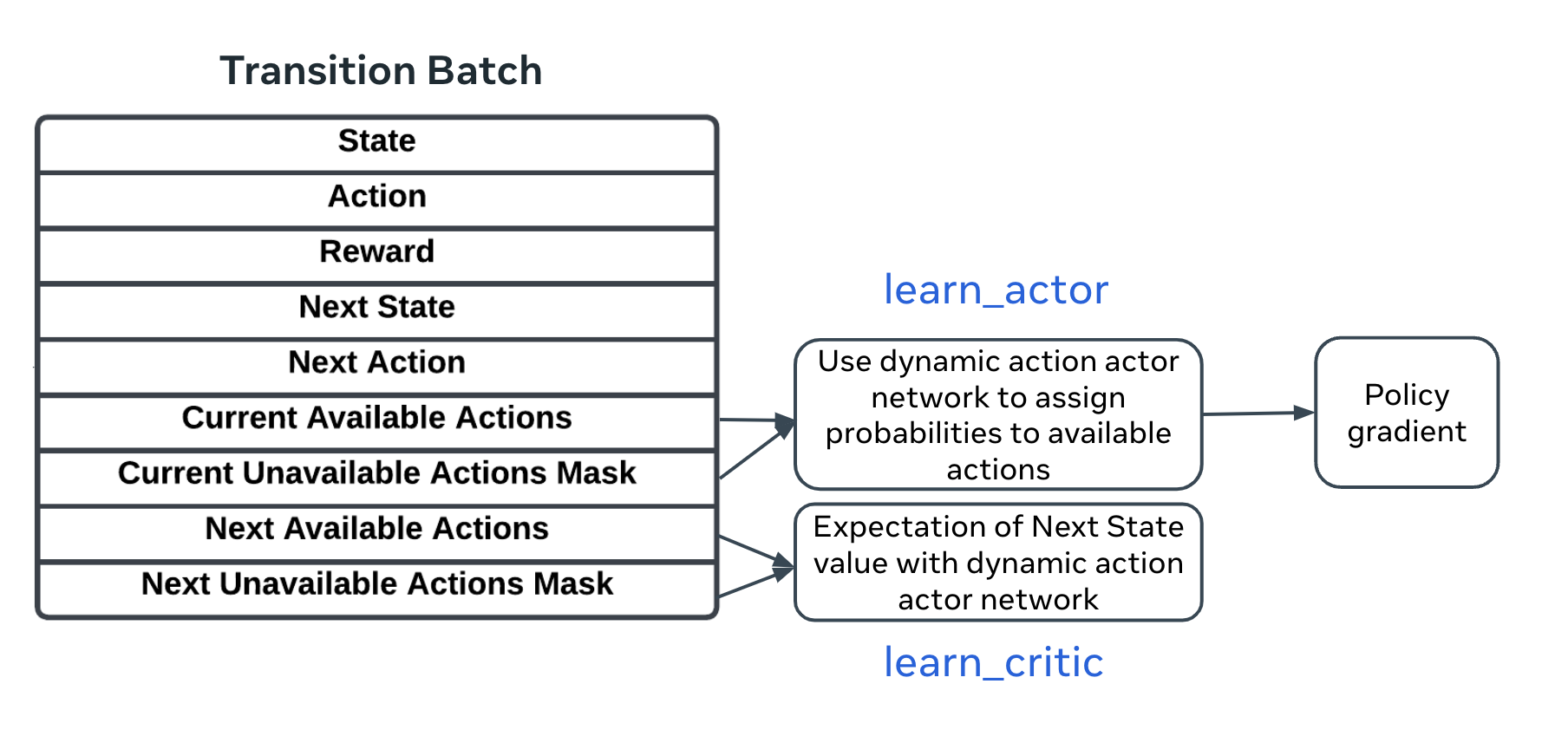}
    \caption{Actor-Critic with Dynamic Action Space Design}
    \label{fig:actor-critic-dynamic-action}
\end{figure}

It is worth noting that the use of dynamic actor networks is critical to our implementation. For critic learning, we compute $\mathbb{E}_{a \sim \pi_\phi}[Q_\theta(S_{t+1}, a)]$ by leveraging the dynamic actor network $\pi_\phi$; here using $\pi_\phi$ ensures that the policy distribution only covers actions in the available action space when computing the expectation over value functions of the next state. Similarly, we also use the dynamic actor network $\pi_\phi$ for policy learning updates. Specifically, in computing the policy gradients, we only use critic values for actions in the available action set of a state. Please refer to Figure \ref{fig:actor-critic-dynamic-action} for an illustration of the changes we make to actor-critic methods for learning with dynamic action spaces.


\section{Additional Details for Industry Adoption Examples}
In this section, we offer the detailed setup of the examples we presented in Section 4. 

\begin{enumerate}
    \item Auction-based recommender systems \citep{xu2023optimizing}:
    \begin{enumerate}
        \item MDP setup: We use RL for a contextual MDP where each user represents a context and the goal is to optimize the cumulative positive feedback throughout users interactions with the recommender systems. 
        \item Observation: User real-time representation is provided by the system summarizing the historical interactions from the user and their real-time privacy-preserving features. We observe feedback from the user after a recommendation is offered to the user. Pearl's history summarization module can also be leveraged if the user time-series representation is not offered by the system.
        \item Action: Each piece of content available for recommendation is an action and the system offers each recommendation to the agent as a representation summarizing the content and the historical interaction outcomes of the content. Since at every step, there is a different set of recommendations, dynamic action space needs to be supported for an RL agent to solve this problem. 
        \item Reward: The user's feedback to the recommendation we offered. 
        \item Policy learner: A key challenge to auction-based recommender systems is that the final deployed policy is heavily influenced by the auction system. In \cite{xu2023optimizing}, we adopted Deep SARSA as the policy learner to perform on-policy policy improvement such that we can account for the auction system's impact on the policy and we are currently further improving the outcomes with off-policy offline learning algorithms such as CQL. We leverage value-based dynamic action space, presented in Section \ref{sec:value-based-dynamic-action} for training and inference.  
    \end{enumerate}
    \item Ads auction bidding \citep{korenkevych2023offline}:
    \begin{enumerate}
        \item MDP setup: RL is leveraged to sequentially place bids in an ads auction system such that the agent can maximize cumulative conversions in an ads campaign. 
        \item Observation: We observe budget consumed, a rough market forecast statistics and the auction bidding results.
        \item Action: The amount of budget the agent places into the ads auction.
        \item Reward: Ads outcome based on a winning auction, otherwise 0. 
        \item Policy learner: We adopt TD3BC as an offline RL algorithm to learn from offline data collected from production bidding systems. 
        \item History summarization: To reduce partial observability of the auction market, we stack some past observations to provide more meaningful insights into current decision making. 
        \item Online exploration: The RL agent leverages Gaussian distribution exploration policy with clipping to offer better support in offline policy learning.  
        \item Safe learning: Our current efforts on auction bidding involve reward-constrained policy optimization (RCPO) in combination with TD3 that prevent the RL agent from taking unsafe actions. 
    \end{enumerate}
    \item Creative selection:
    \begin{enumerate}
        \item Contextual bandit setup: Each context in the creative selection problem is a piece of content to be delivered. 
        \item Observation and reward: In creative selection, the reward and the observation are the same, representing the user's feedback after seeing the piece of content.
        \item Action: Each action is a potential format of the content to be delivered.
        \item Policy learning + Exploration: we use Neural-LinUCB for both learning the expectation of the reward as well as online exploration via the "optimism facing uncertainty" principle. 
    \end{enumerate}
\end{enumerate}

\section{Limitations and Future Work}
When comparing to other libraries, there are three areas where other libraries may offer better support: (1) multi-agent RL, (2) model-based RL and (3) low-level abstractions.
\begin{enumerate}
    \item Multi-agent RL: For our production use cases, we have not yet encountered a significant need to use multi-agent RL algorithms, so our initial release of Pearl focuses on the single-agent setting. However, we do believe that as RL methodology matures in the industry, eventually there will be a need for a production-focused multi-agent RL library (one example of a domain requiring a multi-agent consideration is the setting of ad auctions, where multiple bidders are participating in a single auction). At that point, we believe that Pearl’s agent-focused design can be a great starting point to support such algorithms, and therefore, this is one important potential future direction of Pearl. For now, there are some fantastic specialized libraries for the multi-agent setting (e.g., Mava \citep{de2021mava}, MARLLib \citep{hu2022marllib}, MALib \citep{zhou2023malib}, to name a few).
    
    \item Model-based RL: Model based RL is a growing area of active research for the RL community. As opposed to model free methods, it involves modeling the transition and reward function explicitly rather than learning from environment interaction tuples. We have not yet planned to incorporate model based RL algorithms in Pearl for two reasons: 1) the algorithms in the literature today are exploratory and still not widely adopted in practice although specific applications in robotics are emerging \citep{wu2023daydreamer}, 2) for different production use cases it is typically hard to accurately model the transition and reward functions with only partial information from the environment interactions - for example, in bidding systems, only the winning bid is observed as opposed to all the bids. Moreover, in many systems with human interactions (recommendation systems etc.), the transition and reward functions might be non-stationary, making accurate modeling more challenging. Despite this, there is growing literature on model based RL ideas - with large pre-trained foundation models serving as models of the environment such as Dreamer \citep{hafner2023mastering} and Diffusion World Model \citep{ding2024diffusion}. As these methods mature and start to be adopted by practitioners, we think incorporating them in Pearl will be useful. For now, we encourage users to check out MBRL-Lib \citep{pineda2021mbrl}, an excellent PyTorch based open-source library specialized for implementations of model based RL methods.

    \item Low-level abstractions: Pearl focuses primarily on the design of a comprehensive and production-ready RL agent, so for its initial release, we did not focus on low-level engineering contributions. Some libraries place a strong emphasis on such improvements: for example, one of TorchRL’s \citep{bou2023torchrl} main innovations is the notion of TensorDict, a data container for passing around tensors and other objects between components of an RL algorithm, designed specifically for RL. One potential future extension of Pearl is to make use of such low-level abstractions with the hope of improving performance.

\end{enumerate}

In terms of other future work, a couple of important features we have in mind could be beneficial to the community. First of all, we believe transformer \citep{vaswani2017attention} architectures could offer a great extension to the current history summarization module such that attention can go to past observations that provide meaningful information to the agent's subjective state. Second, our plan in offering support in language model integration with token-level decisions could also benefit researchers interested in the intersection between RL and large language models. 

\section{Additional Details of Algorithms and Selection Criteria}
In this section, we provide a brief introduction for each algorithm implemented in Pearl and discuss when the algorithm should be selected. 

\subsection{Value-Based Sequential Decision Policy Optimization Algorithms}
\begin{itemize}
    \item DQN \citep{mnih2015human}: DQN combines Q-learning with deep neural networks to approximate the optimal action-value function for decision-making in complex environments. It utilizes experience replay and a target network to stabilize training, allowing it to learn effective policies from high-dimensional input spaces and discrete action space. 
    \item Double DQN \citep{van2016deep}: Double DQN addresses the overestimation bias inherent in Q-learning by decoupling the action selection and action evaluation processes. In Double DQN, the action that maximizes the Q-value is selected using the online network, while the value of this action is evaluated using the target network, leading to more accurate and stable learning. This modification improves the performance and robustness of DQN, particularly in environments where overestimation can significantly impact the learning process.
    \item Dueling DQN \citep{wang2016dueling}: Dueling DQN introduces a novel architecture that separates the representation of state values and advantage values within the network. This architecture consists of two streams: one that estimates the state value function and another that estimates the advantage function for each action, which are then combined to form the final Q-value function. This separation allows the network to more effectively learn which states are valuable, independent of the action taken, improving learning efficiency and performance in environments with many similar-valued actions.
    \item Deep SARSA \citep{rummery1994line}: Deep SARSA (State-Action-Reward-State-Action) extends the SARSA method by using deep neural networks to approximate the action-value function. Unlike DQN, which uses the maximum future reward for updating Q-values, Deep SARSA updates its Q-values based on the actual action taken by the policy, incorporating the on-policy nature of SARSA into deep learning. This approach can lead to more stable learning in certain environments, particularly when using stochastic or exploratory policies.
\end{itemize}
\subsection{Actor-critic Sequential Decision Policy Optimization Algorithms}
\begin{itemize}
    \item REINFORCE \citep{sutton1999policy}: REINFORCE is a policy gradient algorithm that directly optimizes the policy by adjusting its parameters to maximize the expected reward. It uses the gradient of the log-probability of the chosen actions, weighted by the cumulative reward, to update the policy parameters, allowing it to learn complex policies in high-dimensional spaces. While effective, REINFORCE often suffers from high variance in the gradient estimates, making it necessary to use techniques like baselines to reduce variance and improve learning stability.
    \item DDPG \citep{silver2014deterministic}: DDPG is a model-free, off-policy actor-critic algorithm designed for continuous action spaces. It combines the benefits of deterministic policy gradients, which reduce variance in gradient estimates, with deep neural networks to learn both the policy (actor) and value function (critic). DDPG employs experience replay and uses separate target networks for stable training, enabling it to perform well in complex environments with high-dimensional state and action spaces.
    \item TD3 \citep{fujimoto2018addressing}: TD3 addresses key limitations by incorporating three critical improvements: clipped double Q-learning, delayed policy updates, and target policy smoothing. Clipped double Q-learning mitigates overestimation bias by using the minimum of two Q-value estimates, while delayed policy updates slow down the policy changes to provide more accurate value estimates. Target policy smoothing adds noise to target actions to reduce overfitting to narrow peaks in the value function, resulting in more robust and stable learning in continuous action spaces.
    \item SAC \citep{haarnoja2018soft}: SAC is an off-policy algorithm that aims to maximize both the expected reward and the entropy of the policy, promoting exploration and robustness. It utilizes a stochastic policy, represented by a Gaussian distribution, and employs an entropy term in the objective function to encourage diverse and exploratory actions. SAC combines the benefits of both actor-critic and entropy-regularized methods, resulting in stable and efficient learning in high-dimensional continuous action spaces. We also implemented the discrete action space version by computing expected value function and policy distribution entropy from a discrete policy distribution.
    \item PPO \citep{schulman2017proximal}: PPO strikes a balance between the stability of trust-region methods and the efficiency of first-order optimization techniques. It introduces a clipped surrogate objective to limit the policy updates, ensuring that changes to the policy are not too drastic and remain within a predefined trust region. This approach simplifies implementation while maintaining robust performance and sample efficiency, making PPO a popular choice for a wide range of RL tasks.
\end{itemize}

\subsection{Exploration Algorithms}
\begin{itemize}
    \item LinUCB \citep{li2010contextual}: LinUCB selects actions based on upper confidence bounds derived from linear models. It balances exploration and exploitation by considering both the predicted reward and the uncertainty of the prediction, allowing it to make informed decisions in dynamic environments. LinUCB is particularly effective in scenarios where the reward is a linear function of known features, making it a powerful tool for personalized recommendations and adaptive decision-making.
    \item Neural-LinUCB \citep{xu2021neural}: Neural-LinUCB enhances the traditional LinUCB algorithm by integrating deep neural networks to capture complex, non-linear relationships in the contextual data. In this hybrid approach, a neural network extracts high-dimensional feature representations, which are then fed into a linear model to compute the upper confidence bounds for action selection. This combination allows Neural-LinUCB to leverage the expressive power of neural networks for more accurate predictions while maintaining the principled exploration-exploitation balance of LinUCB, making it effective for sophisticated decision-making tasks.
    \item LinTS \citep{agrawal2013thompson}: LinTS leverages Bayesian inference to balance exploration and exploitation by maintaining a probability distribution over the model parameters. For each action, it samples parameters from this distribution to estimate the reward, effectively incorporating uncertainty in the decision-making process. This approach allows the algorithm to adaptively learn the optimal policy in environments where the reward is a linear function of the context, making it suitable for applications such as online recommendations and personalized marketing.
    \item SquareCB \citep{foster2020beyond}: SquareCB addresses the exploration-exploitation trade-off in contextual bandit problems by reducing them to online regression tasks. It utilizes an online regression oracle with squared loss to efficiently manage and update confidence bounds for decision-making. This approach allows SquareCB to optimize performance under both agnostic and realizable settings, providing a robust solution for contextual bandits in diverse applications.
    \item Bootstrapped DQN and Randomized Value Functions \citep{osband2016deep}: Bootstrapped DQN enhances the exploration capabilities of the standard Deep Q-Network (DQN) by introducing multiple heads in the output layer of the neural network, each representing a different Q-function. These heads are trained on slightly different subsets of the training data, effectively creating diverse value estimations that encourage exploration in the learning process. This approach addresses the challenge of insufficient exploration in standard DQN, improving the algorithm's performance in environments with sparse rewards.
\end{itemize}

\subsection{Offline Learning Algorithms}
\begin{itemize}
    \item CQL \citep{kumar2020conservative}: CQL is designed to effectively learn from offline data without further interaction with the environment. CQL introduces a regularization term that penalizes the Q-values of unseen actions, reducing overestimation and promoting conservative policy estimates. This approach leads to more stable and robust policy evaluation, especially in settings where the available data is limited or biased.
    \item IQL \citep{kostrikov2021offline}: IQL aims to learn optimal policies from a fixed dataset without further environment interaction. It uses a behavior cloning loss for action selection, combined with a standard Q-learning loss for policy evaluation, which separates the action selection from policy evaluation to stabilize training. This approach helps in handling the distributional shift between the policy learned from offline data and the optimal policy, making IQL effective in environments where exploration is limited.
    \item TD3BC \citep{fujimoto2021minimalist}: TD3BC combines the TD3 algorithm with behavior cloning to address the distributional shift challenge in offline RL. It leverages the stability and exploration-exploitation balance of TD3 while using behavior cloning to constrain policy updates closer to the data distribution. This integration enhances sample efficiency and improves performance by reducing the detrimental effects of extrapolation errors common in offline settings.
\end{itemize}

\subsection{Safe Learning Algorithms}
\begin{itemize}
    \item Risk-sensitive learning via QRDQN \citep{dabney2018distributional}: QRDQN approximates the full distribution of returns instead of just their mean by using quantile regression. This method represents the expected return distribution using multiple quantiles, which allows the policy to better estimate the variability and risk associated with different actions. By adjusting the quantiles to emphasize lower parts of the distribution, QRDQN can be tailored to prioritize risk-averse behaviors, producing policies that prefer actions leading to lower variability in received rewards, thus enabling risk-sensitive learning.
    \item Reward constrained policy optimization (RCPO) \citep{tessler2018reward}: RCPO is an approach for solving constrained policy optimization problems. It integrates penalty signals into the reward function to guide policy towards satisfying constraints, while also proving convergence under mild assumptions. RCPO is particularly effective in robotic domains, demonstrating faster convergence and improved stability compared to traditional methods, making it adept at handling both general and specific constraints within RL tasks. We have made RCPO compatible with TD3 \citep{fujimoto2018addressing}, SAC \citep{haarnoja2018soft} and DDPG \citep{silver2014deterministic}. 
\end{itemize}

\subsection{History Summarization Algorithms}
Our main approaches for history summarization are history observation stacking and LSTM \citep{hochreiter1997long} summarization. 
\begin{itemize}
    \item Stacking observations: Stacking observations in RL, especially in partially observable environments, provides a practical way to approximate a fully observable state. By incorporating a sequence of the most recent observations, actions, or both, agents can infer the underlying state of the system more accurately. This method effectively turns the problem into one that resembles a fully observable Markov decision process (MDP), enabling the agent to make more informed decisions based on the extended context provided by the historical data.
    \item LSTM history summarization: Using an LSTM (long short-term memory) network in RL, particularly in partially observable environments, enhances the agent's ability to remember and utilize long-term dependencies in the observed data \citep{lambrechts2022recurrent}. This capability is crucial where stacking history alone might not capture the necessary context for decision-making. LSTMs help process sequences of past observations, maintaining a more dynamic and informative state representation, which allows the agent to make more accurate predictions and decisions based on the extended sequence of past events. This approach can significantly improve performance in environments where the agent's current state is not fully observable through immediate inputs alone.
\end{itemize}

\newpage
\bibliography{bibliography}

\end{document}